\documentclass[pdflatex, sn-basic, iicol, Numbered]{sn-jnl}

\usepackage{graphicx}%
\usepackage{xcolor}%
\usepackage{multirow}%
\usepackage{amsmath,amssymb,amsfonts}%
\usepackage{amsthm}%
\usepackage{mathrsfs}%
\usepackage[title]{appendix}%
\usepackage{textcomp}%
\usepackage{manyfoot}%
\usepackage{booktabs}%
\usepackage{algorithmicx}%
\usepackage{algpseudocode}%
\usepackage{listings}%
\usepackage{nicematrix}
\usepackage{pgfgantt}
\usepackage{geometry}
\usepackage{lscape}
\usepackage{adjustbox}
\usepackage{tabularx}
\usepackage{booktabs}
\usepackage{makecell}
\usepackage{array}
\usepackage{soul}
\usepackage{tikz}
\usepackage{caption}
\usepackage{subcaption}
\usepackage{float}
\usepackage{booktabs}

\theoremstyle{thmstyleone}%
\theoremstyle{thmstyletwo}%

\theoremstyle{thmstylethree}%

\begin{document}

\title[Article Title]{Padé Approximant Neural Networks for Enhanced Electric Motor Fault Diagnosis Using Vibration and Acoustic Data}


\author*[1,2]{\fnm{Sertac} \sur{Kilickaya}}\email{sertac.kilickaya@ieu.edu.tr}

\author[1]{\fnm{Levent} \sur{Eren}}\email{levent.eren@ieu.edu.tr}

\affil[1]{\orgdiv{Department of Electrical and Electronics Engineering}, \orgname{Izmir University of Economics}, \orgaddress{\street{Sakarya Street No:156}, \city{Izmir}, \postcode{35330}, \country{Turkey}}}

\affil[2]{\orgdiv{Faculty of Information Technology and Communication Sciences}, \orgname{Tampere University}, \orgaddress{\street{Korkeakoulunkatu 7}, \city{Tampere}, \postcode{33720}, \country{Finland}}}


\abstract{
\textbf{Purpose} The primary aim of this study is to enhance fault diagnosis in induction machines by leveraging the Padé Approximant Neuron (PAON) model. While accelerometers and microphones are standard in motor condition monitoring, deep learning models with nonlinear neuron architectures offer promising improvements in diagnostic performance. This research investigates whether Padé Approximant Neural Networks (PadéNets) can outperform conventional Convolutional Neural Networks (CNNs) and Self-Organized Operational Neural Networks (Self-ONNs) in the diagnosis of electrical and mechanical faults from vibration and acoustic data.

\textbf{Methods} We evaluate and compare the diagnostic capabilities of three deep learning architectures: one-dimensional CNNs, Self-ONNs, and PadéNets. These models are tested on the University of Ottawa's publicly available constant-speed induction motor datasets, which include both vibration and acoustic sensor data. The PadéNet model is designed to introduce enhanced nonlinearity and is compatible with unbounded activation functions such as LeakyReLU.

\textbf{Results and Conclusion} PadéNets consistently outperformed the baseline models, achieving diagnostic accuracies of 99.96\%, 98.26\%, 97.61\%, and 98.33\% for accelerometers 1, 2, 3, and the acoustic sensor, respectively. The enhanced nonlinearity of PadéNets, together with their compatibility with unbounded activation functions, significantly improves fault diagnosis performance in induction motor condition monitoring.
}




\maketitle

\section{Introduction}\label{sec1}

Electrical machines are the backbone of modern industrial processes, driving manufacturing, automation and energy production. However, continuous operation over time leads to inevitable wear and an increased risk of failure \cite{kumar2022comprehensive}. To address this, condition monitoring has emerged as an essential practice, employing traditional model-based, signal-based, and modern data-driven approaches to evaluate the health of motors, generators, and other rotating equipment. Model-based methods rely on physical or mathematical representations of machine behavior, signal-based techniques use signal processing to analyze sensor outputs such as vibration and sound, and data-driven models harness artificial intelligence (AI) to uncover and detect patterns from raw data \cite{ince2016real}. Ensuring reliability through these strategies minimizes costly downtimes and prevents safety hazards.

Beyond diagnostic techniques, advancements in sensor technologies have been instrumental in improving fault diagnosis capabilities, providing richer, more precise data to feed these monitoring systems. Modern industrial systems are often monitored using a variety of sensors that track different parameters such as temperature, current, sound, vibration, and visual data like images or videos. Accelerometers are the most commonly used sensors for machinery fault diagnosis, favored for their high sensitivity, high dynamic range, and wide bandwidth in frequency response \cite{yan2023review, abid2021review}. During rotary motion, components of rotating machines produce vibrations, with characteristic frequencies determined by their rotational speed, geometry, and interactions with other parts \cite{ince2016real}. The vibration amplitude at a specific frequency is predictable but increases with wear or damage, making Fast Fourier Transform (FFT) analysis crucial for detecting fault-induced changes \cite{bogue2013sensors}. However, accelerometers are contact sensors, and their response can significantly vary depending on the mounting location, which is one of the most common problems associated with these sensors. As a non-contact alternative, microphones can be used for condition monitoring, offering several advantages such as lower cost, easier installation, and the ability to monitor multiple machines simultaneously without the need for physical attachment \cite{tang2023survey, kilickaya2024audio}. Acoustic-based monitoring eliminates the need for mechanical coupling and avoids potential sensor mounting resonances that can distort vibration measurements. These diverse sensing modalities generate complex, high-dimensional time series data often at high sampling rates, resulting in vast amounts of information in industrial environments. Deep learning (DL) models can handle high-velocity data streams more effectively than traditional spectral analysis and statistical pattern recognition methods for several reasons. They process high-dimensional data, automatically extract relevant features end-to-end without manual intervention \cite{ince2016real, kumar2022comprehensive}, capture complex, nonlinear relationships, and adapt to evolving data over time \cite{lecun2015deep, celik2021adaptation}. Additionally, DL architectures scale efficiently, managing vast datasets from multiple sensors in industrial environments \cite{kibrete2024multi, qiu2023deep, yang2024multisensor}. Consequently, DL architectures, such as Convolutional Neural Networks (CNNs) \cite{ince2016real, janssens2016convolutional, jiao2020comprehensive, celebioglu2024smartphone}, Recurrent Neural Networks (RNNs) \cite{zhu2022application, zhang2021fault, yuan2016fault}, and hybrid models \cite{borre2023machine, guo2023rolling, xiang2021fault, ertargin2024mechanical}, potentially incorporating attention mechanisms, have become widely adopted computational frameworks for fault diagnostics.

CNNs have been widely adopted for machine fault classification, which is one of the earliest and most extensively studied applications of fault diagnosis, drawing direct inspiration from image classification techniques \cite{neupane2025data}. In this context, both one-dimensional (1D) and two-dimensional (2D) CNNs have been utilized. The 1D CNNs are specialized for processing time series data such as raw vibration or audio signals, whereas the 2D CNNs are designed to manage multidimensional data, often by converting 1D signals into 2D representations that capture both spatial and temporal relationships. For example, in \cite{youcef2020rolling}, vibration spectrum imaging (VSI) was used to transform normalized spectral amplitudes from segmented vibration signals into images. These images were subsequently fed into a CNN for bearing fault classification. The proposed VSI-CNN model achieved a classification accuracy of around 99\%. Similarly, in \cite{neupane2020deep}, a 2D CNN model achieved an accuracy of 99.38\% by utilizing 2D image representations of 1D raw vibration data from the Case Western Reserve University (CWRU) bearing dataset. In addition to these transformed inputs, thermal images have also been utilized as inputs for 2D CNNs in fault diagnosis \cite{yongbo2020rotating, piechocki2023unraveling}. On the other hand, 1D CNNs provide a simple and computationally efficient way to perform fault diagnosis by directly processing raw 1D input data. Numerous studies \cite{ince2016real, chen2020improved, celebioglu2024smartphone, zhang2018deep, eren2017bearing} have applied 1D CNNs to machinery fault diagnosis, using either raw sensor data or engineered features as input. 

While CNNs exhibit strong performance under controlled conditions, prior studies \cite{kiranyaz2017progressive, kiranyaz2020operational} emphasize that conventional CNNs, built upon a fixed architecture and first-order neuron model, often struggle to capture highly nonlinear and complex patterns inherent in real-world data. Although nonlinearity is introduced through pointwise activation functions, such as ReLU \cite{nair2010rectified} and its variant $\mathrm{LeakyReLU}$ \cite{maas2013rectifier}, these functions are predefined and uniformly applied across layers, limiting the network's representational flexibility. To address this constraint, Padé Activation Units (PAUs) were introduced in \cite{molina2019pad} as a learnable alternative to traditional hand-crafted activations. PAUs model activation functions using Padé approximants, which are rational functions expressed as the ratio of two polynomials. This formulation enables the network to learn complex, task-specific nonlinear mappings during training, as both the numerator and denominator coefficients are optimized via backpropagation (BP). By making the activation functions adaptive rather than fixed, PAUs provide a more expressive and flexible framework for capturing intricate data patterns. To further extend this paradigm and enhance the network's ability to model nonlinearities at the neuron level, Self-Organized Operational Neural Networks (Self-ONNs) have been proposed \cite{kiranyaz2021self}. Unlike traditional CNNs, which rely solely on pointwise nonlinear activations, Self-ONNs incorporate nonlinear neuron models. They incorporate generative neurons that approximate the necessary nonlinear mappings by utilizing a truncated Taylor series expansion centered at the origin, specifically applying a Maclaurin series expansion up to a predefined order. Therefore, generative neurons use the input along with its higher-order powers and compute their weighted sum to approximate a nonlinear mapping in the neuron itself. The enhanced fault diagnosis performance of 1D and 2D Self-ONNs has been validated in studies on machinery fault diagnosis, utilizing various sensor modalities \cite{ince2021early, ince2022improved, kilickaya2024bearing, kilickaya2024thermal}. While generative neurons in Self-ONNs capture greater nonlinearity, the linear combination of different input orders can lead to instability outside a safe computation range. Moreover, since Taylor series approximations are most accurate near the expansion point, the output of generative neurons is constrained by a $\tanh$ activation function, which may suffer from vanishing gradients during training. To address this limitation, a new class of networks known as PadéNets, built using Padé Approximant Neurons (PAONs) has been proposed \cite{Pade}. Padé neurons utilize Padé approximation at the neuron level by representing nonlinear functions as ratios of polynomials. In PadéNets, a single neuron within a 1D Padé layer with a kernel size of $k$ effectively learns $k$ distinct Padé approximants, each represented as a ratio of two polynomials. This structure significantly increases the degrees of freedom available to the model compared to PAUs, as it introduces nonlinearity within the kernel itself, in addition to the nonlinearity contributed by the activation function. They have been shown to offer improved performance compared to Taylor-based generative neurons and convolutional neurons in single-image super-resolution tasks \cite{Pade}. Furthermore, PAONs generalize several existing neuron models and can effectively serve as a replacement for conventional convolutional neurons within CNN architectures. To leverage the superior feature extraction capabilities of PAONs, this study introduces the use of 1D PadéNets for the classification of electrical and mechanical faults in three-phase induction machines. The main contributions of this work can be summarized as follows:

\begin{itemize} 
\item We present, for the first time, the application of 1D PadéNets for the classification of electrical and mechanical faults in three-phase induction motors.
\item We evaluate 1D PadéNets separately on benchmark vibration and audio datasets from the University of Ottawa \cite{sehri2024university}, demonstrating robust results across different sensing modalities. 
\item We compare 1D PadéNets with 1D Self-ONNs and 1D CNNs to highlight the superior diagnostic accuracy enabled by Padé neurons.
\end{itemize}

\begin{figure*}[htbp]
\includegraphics[width=\textwidth]{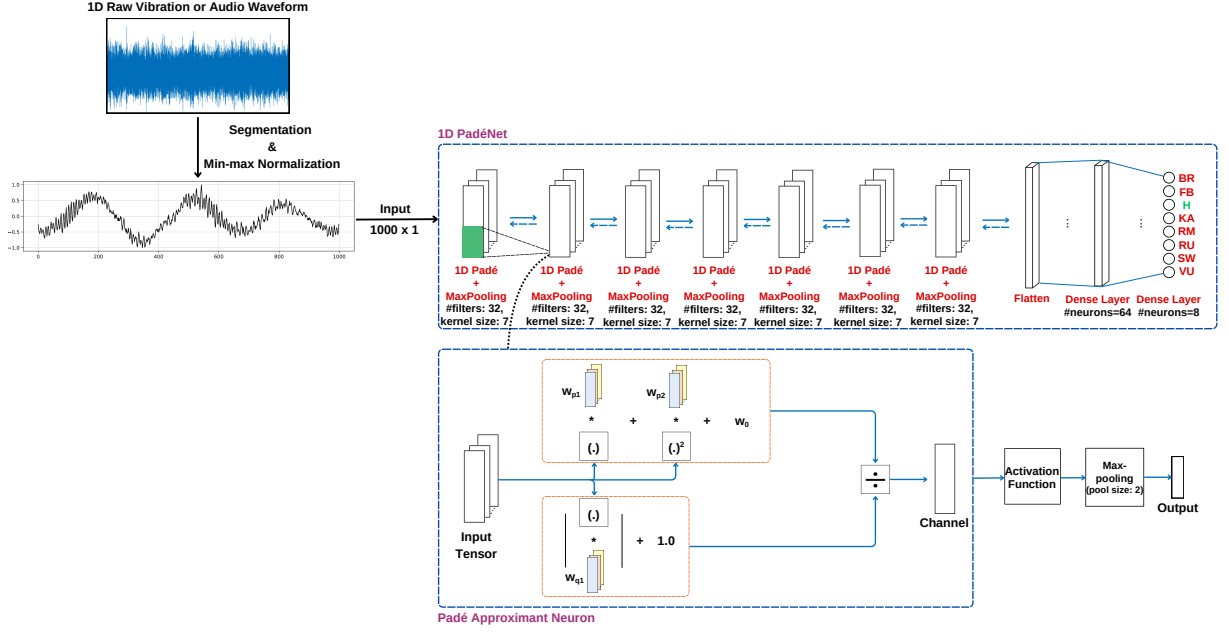}
\caption{The proposed 1D PadéNet-based framework and the diagram of a Padé neuron with $P = 2, Q = 1$, where $w_{0}$ represents the bias term in the numerator, $(\cdot)^{n}$ denotes element-wise exponentiation of the input to the $n^{\text{th}}$ power, $\mathbf{\ast}$ indicates convolution, and $\frac{\displaystyle\cdot}{\displaystyle\cdot}$ implements Equation~\ref{eq9}. \label{fig1}}
\end{figure*}

The proposed 1D PadéNet-based framework for fault diagnosis is illustrated in Figure~\ref{fig1}. Section~\ref{sec2} discusses the mathematical foundations of all evaluated models: 1D CNNs, 1D Self-ONNs, and 1D PadéNets. Section~\ref{sec3} presents the experimental setup, including the University of Ottawa constant-speed vibration and acoustic datasets \cite{sehri2024university}, followed by details on preprocessing steps, training methodology, and evaluation metrics. Section~\ref{sec4} presents a thorough comparison of the fault diagnosis performance of each model, along with their computational complexities. Finally, Section~\ref{sec5} concludes the paper and outlines potential directions for future research.

\section{Methods} \label{sec2}
This section establishes the mathematical foundations and architectural characteristics of 1D CNNs, 1D Self-ONNs, and 1D PadéNets to enable their comparative evaluation in electric motor fault diagnosis.

1D CNNs achieved state-of-the-art performance in various applications, including biomedical data classification \citep{kiranyaz2015real}, structural health monitoring \citep{abdeljaber2017real}, and motor fault diagnosis \citep{ince2016real, chen2020improved, celebioglu2024smartphone, zhang2018deep, eren2017bearing}. Their simple 1D convolutional structure also enables real-time, low-cost hardware implementation \citep{kiranyaz20211d}. The traditional CNN architecture is based on the classical linear neuron model, which incorporates constraints such as restricted connectivity and weight sharing at the kernel level. These constraints lead to the convolution operations commonly used in CNNs. The \( k^\text{th} \) input feature map in the \( l^\text{th} \) layer of a 1D CNN can be computed as:
\begin{equation}
\mathbf{x}_k^{(l)} 
= b_k^{(l)} 
+ \sum_{i=1}^{N_{l-1}} \mathbf{x}_{ik}^{(l)} 
\label{eq1}
\end{equation}

In this expression, $\mathbf{x}_{ik}^{(l)} \in \mathbb{R}^M$ denotes the feature map obtained by convolving the $i^\text{th}$ output map from layer $(l-1)$, denoted as $\mathbf{y}_i^{(l-1)} \in \mathbb{R}^M$, with the kernel $\mathbf{w}_{ik}^{(l)} \in \mathbb{R}^K$, which connects it to the $k^\text{th}$ input feature map in layer $l$. The term $b_k^{(l)}$ represents the bias associated with the $k^\text{th}$ neuron in the current layer, and $N_{l-1}$ is the number of output feature maps (or channels) produced by layer $l-1$. The 1D convolution operation used to compute $\mathbf{x}_{ik}^{(l)}[m]$ is given by:
\begin{equation}
\mathbf{x}_{ik}^{(l)}[m] 
= \sum_{r=0}^{K-1} w_{ik}^{(l)}[r] \, y_i^{(l-1)}[m+r]
\label{eq2}
\end{equation}
In the forward pass, each input feature map $\mathbf{x}_{k}^{(l)}$ is then transformed by a nonlinear activation function followed by an optional subsampling operation, resulting in the output feature representation of the convolutional neuron.

CNNs are derived from the traditional McCulloch-Pitts neuron model, which is fundamentally linear, with nonlinearity introduced through an activation function. To extend nonlinearity beyond simple pointwise transformations, new architectures like Operational Neural Networks (ONNs) \citep{kiranyaz2020operational}, which incorporate inherent nonlinearities within their neurons, have been proposed. ONNs extend the conventional convolutional neuron by generalizing the standard convolution operation as follows:
\begin{equation}
\overline{\mathbf{x}}_{ik}^{(l)}[m] 
= P_k^{(l)} \!\left( \psi_k^{(l)}\!\left(w_{ik}^{(l)}[r],\; y_i^{(l-1)}[m+r]\right)_{r=0}^{K-1} \right)
\label{eq3}
\end{equation}
where $\boldsymbol{\psi}_{k}^{(l)}(\cdot): \mathbb{R}^{M \times K} \rightarrow \mathbb{R}^{M \times K}$ and $P_{k}^{(l)}(\cdot): \mathbb{R}^K \rightarrow \mathbb{R}^1$ are called nodal and pool operators, respectively, assigned to the $k^\text{th}$ neuron of the $l^\text{th}$ layer.

Operational layers in ONNs preserve the two fundamental constraints of conventional CNNs, namely weight sharing and localized connectivity at the kernel level. However, they can utilize a variety of functions as the nodal operators, including sinusoidal transformations, exponentials, or other nonlinear operations \citep{kiranyaz2020operational}. Additionally, instead of the standard additive pooling in CNNs, these models allow for alternative aggregation strategies such as taking the median, offering greater flexibility in learning complex patterns. In ONNs, the Greedy Iterative Search (GIS) algorithm is often employed to explore a set of candidate functions, aiming to determine the most effective combination of nodal and pooling operators \citep{kiranyaz2020operational}. Once these optimal operators are selected, they are uniformly assigned to all neurons within a given hidden layer, defining the final structure of the network. Despite its effectiveness, this design introduces key limitations \citep{kiranyaz2021self}. A major drawback is the lack of diversity, as each neuron within a layer uses the same operator set, limiting functional heterogeneity. Additionally, identifying appropriate candidate operators prior to training poses a significant challenge, because it requires considerable computational effort and may introduce bias that affects learning. To overcome these issues, Self-ONNs were introduced \citep{kiranyaz2021self}. 

Self-ONNs leverage a generative neuron model to enable adaptive operator selection during training. Each generative neuron can optimize its nodal operators through BP training. This optimization occurs individually for each kernel element and connection to neurons in the previous layer, with the goal of maximizing learning performance. In self-organized operational layers, the nodal functions are optimized by approximating nonlinear behaviors using a Taylor series expansion. This approach enables each generative neuron to apply a learned nodal transformation, which can be formulated as follows:
\begin{equation}
\begin{aligned}
&\tilde{\boldsymbol{\psi}}_{k}^{(l)}\!\left( \{\mathbf{w}_{p,ik}^{(l)}[r]\}_{p=1}^P,\, \mathbf{y}_i^{(l-1)}[m+r] \right) \\
&\quad= \sum_{p=1}^{P} \mathbf{w}_{p,ik}^{(l)}[r] \; \big( \mathbf{y}_i^{(l-1)}[m+r] \big)^{p}
\end{aligned}
\label{eq4}
\end{equation}

In Equation~\ref{eq4}, the hyperparameter $P$ sets the order of the Taylor polynomial approximation, thereby influencing the degree of nonlinearity. Additionally, the weights $\mathbf{w}_{p,ik}^{(l)}$ now consist of $P$ times the number of learnable parameters in the corresponding convolutional model. During training, the weights $\mathbf{w}_{p,ik}^{(l)}$ are updated via the standard BP algorithm, leading to nonlinear transformations \citep{kiranyaz2021self}.

By adopting summation as the pooling operator, we can model the self-organized operational layer using a convolutional framework. The output of a generative neuron can simply be expressed as follows:
\begin{equation}
\tilde{\mathbf{x}}_{ik}^{(l)} 
= \sum_{p=1}^{P} \operatorname{Conv1D}\!\left( \mathbf{w}_{p,ik}^{(l)},\; \big( \mathbf{y}_i^{(l-1)} \big)^{p} \right) 
\label{eq5}
\end{equation}

Thus, the formulation can be implemented using $P$ 1D convolution operations. When $P = 1$, it simplifies to the conventional 1D convolution. Self-ONNs are super set of CNNs ($P = 1$), and generative neurons in Self-ONNs enable the modeling of more complex nonlinearities. However, using higher-order powers $\big(\mathbf{y}_{i}^{(l-1)}\big)^{p}$ can introduce numerical instability outside a stable computational range; therefore, $\tilde{\mathbf{x}}_{i k}^{(l)}$ is typically constrained by bounded activation functions. Since Taylor series approximations are most accurate near the expansion point, the output of generative neurons is typically constrained by the $\tanh$ activation function to model the nonlinear mapping around the origin. However, the $\tanh$ function saturates, which can lead to the vanishing gradient problem during BP, hindering effective training. To address these challenges, a new neuron model, Padé Approximant Neurons (PAONs), inspired by Padé approximants, has recently been proposed \citep{Pade}.

\begin{figure*}[htbp]
\includegraphics[width=\textwidth]{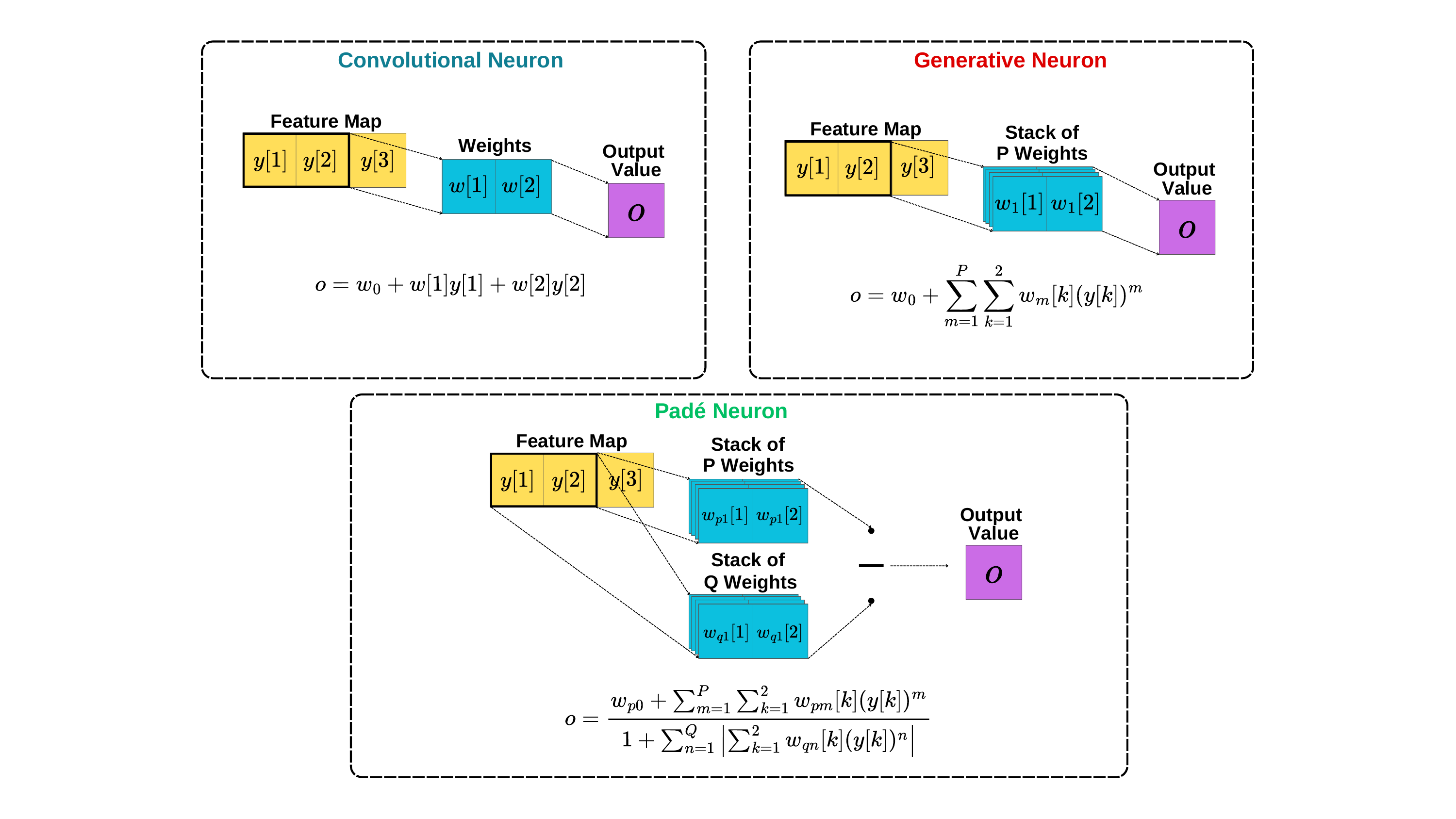}
\caption{An illustrative overview of the operations involved in a convolutional, generative, and Padé neuron. \label{fig2}}
\end{figure*} 

The Padé approximation offers a powerful means of representing transcendental functions by expressing them as a ratio of two polynomials of specified degrees. It finds extensive application in fields such as control theory, where it is particularly useful for approximating time-delay elements in feedback control systems. An asymptotic expansion, such as a Taylor series, can often be significantly accelerated or even transformed from divergent to convergent by reformulating it as a Padé approximant \citep{george1975essentials}.

If we let \( f_{[P/Q]}(y) \) denote the Padé approximation of a function \( f(y) \), where the numerator is a polynomial of degree \( P \) and the denominator is a polynomial of degree \( Q \), the approximation can be formulated as follows:
\begin{equation}
f_{[P/Q]}(y) 
= \frac{R_P(y)}{S_Q(y)} 
= \frac{\displaystyle \sum_{m=0}^{P} a_m\, y^m}{\displaystyle \sum_{n=0}^{Q} b_n\, y^n}
\label{eq6}
\end{equation}

The coefficients \( a_m \) and \( b_n \) correspond to the terms in the numerator and denominator polynomials, respectively. Typically, to simplify the formulation, the Padé approximant coefficients are normalized such that \( b_0 = 1 \). Hence, we can express it as follows:
\begin{equation}
f_{[P/Q]}(y) 
= \frac{\displaystyle a_0 + \sum_{m=1}^{P} a_m\, y^m}{\displaystyle 1 + \sum_{n=1}^{Q} b_n\, y^n}
\label{eq7}
\end{equation}

Padé Activation Units (PAUs) adapt the classical Padé approximation for use as learnable activation functions in neural networks \citep{molina2019pad}. Unlike fixed-form nonlinearities such as $\tanh$, a PAU represents the activation as the ratio of two polynomials whose coefficients are trainable parameters. By appropriately learning these coefficients, PAUs can replicate common activation functions (e.g., sigmoid, $\tanh$) as special cases or generate entirely new, data-driven nonlinearities. The analytical differentiability of the rational form ensures compatibility with standard BP, while its flexibility enables the network to capture complex, task-specific behaviors that may be inaccessible to conventional fixed activations. Formally, a PAU in the \( l^\text{th} \) layer, applied element-wise to each input map \( \mathbf{x}_i^{(l)} \in \mathbb{R}^M \), is defined as:
\begin{equation}
\boldsymbol{\phi}_{\text{PAU}}\!\left(\mathbf{x}_i^{(l)}\right) 
= \frac{\displaystyle \sum_{m=0}^{P} a_{m}^{(l)} \, \big( \mathbf{x}_i^{(l)} \big)^{m}}
       {\displaystyle 1 + \sum_{n=1}^{Q} b_{n}^{(l)} \, \big( \mathbf{x}_i^{(l)} \big)^{n}},
\label{eq_pau}
\end{equation}
where \( a_{m}^{(l)} \) and \( b_{n}^{(l)} \) are trainable scalar coefficients shared by all feature maps in the \( l^\text{th} \) layer.

This idea of embedding trainable Padé approximants into the network structure naturally motivates the use of PAONs in PadéNets, where the polynomial ratio formulation is incorporated directly into the neuron instead of the activation function. If we interpret the coefficients \( a_m \) and \( b_n \) as kernels in a convolution operation, with \( a_0 \) representing the bias, the \( k^\text{th} \) input feature map in the \( l^\text{th} \) Padé layer can be expressed as:
\begin{equation}
\mathbf{x}_k^{(l)} =
\frac{
    w_{p0,k}^{(l)}
    + \displaystyle\sum_{m=1}^{P} \sum_{i=1}^{N_{l-1}}
      \mathbf{w}_{pm,ik}^{(l)} \ast \big( \mathbf{y}_i^{(l-1)} \big)^{m}
}{
    1 + \displaystyle\sum_{n=1}^{Q} \sum_{i=1}^{N_{l-1}}
      \mathbf{w}_{qn,ik}^{(l)} \ast \big( \mathbf{y}_i^{(l-1)} \big)^{n}
}
\label{eq8}
\end{equation}
where $\mathbf{y}_i^{(l-1)} \in \mathbb{R}^M$ is the $i^\text{th}$ output feature map from the $(l-1)^\text{th}$ layer, $\mathbf{w}_{pm,ik}^{(l)},\, \mathbf{w}_{qn,ik}^{(l)} \in \mathbb{R}^K$ are the numerator and denominator kernels corresponding to polynomial orders $m$ and $n$, respectively, with $i$ indexing the output feature map from layer $l-1$ and $k$ indexing the input feature map in layer $l$. $w_{p0,k}^{(l)}$ is the bias term, and $\ast$ denotes 1D convolution. One important consideration with this neuron model is that the denominator can potentially become zero or approach to zero throughout training. To mathematically ensure that the denominator remains nonzero, several variants of the Padé neurons have been proposed \citep{Pade}. In this study, we adopt the first variant, which involves taking the absolute value of each term in the denominator to guarantee that each element in the numerator is divided by a value greater than or equal to one. Therefore, we can express the equation for this variant as follows:
\begin{equation}
\mathbf{x}_k^{(l)} =
\frac{
    w_{p0,k}^{(l)}
    + \displaystyle\sum_{m=1}^{P} \sum_{i=1}^{N_{l-1}}
      \mathbf{w}_{pm,ik}^{(l)} \ast \big( \mathbf{y}_i^{(l-1)} \big)^{m}
}{
    1 + \displaystyle\sum_{n=1}^{Q} \sum_{i=1}^{N_{l-1}}
      \big|\, \mathbf{w}_{qn,ik}^{(l)} \ast \big( \mathbf{y}_i^{(l-1)} \big)^{n} \,\big|
}
\label{eq9}
\end{equation}

Each kernel element in a Padé neuron adapts independently, allowing each weight group to learn its own specific Padé approximation. This self-adjustment enhances the model's nonlinearity by integrating higher-order features in both the numerator and denominator of the approximation. Additionally, as the Padé neuron is expressed as a ratio of polynomials, it provides greater stability, even with higher-order approximations. When the numerator and denominator degrees are comparable, the PAON’s rational form maintains training stability even with unbounded activation functions.

Padé neurons generalize both convolutional and generative neuron models. For \( P = 1 \) and \( Q = 0 \), the Padé neuron reduces to a standard convolutional neuron in CNNs, and for \( P \geq 2 \) and \( Q = 0 \), they behave as generative neurons in Self-ONNs. As a result, PAONs can effectively capture complex nonlinear relationships and they are capable of replacing existing neuron models in CNNs and Self-ONNs. Compared to a standard Conv1D layer, the PAON formulation introduces \((P+Q-1) \times K \times C_{\text{in}} \times C_{\text{out}}\) additional trainable parameters for a kernel size \(K\), where \(C_{\text{in}}\) and \(C_{\text{out}}\) denote the number of input and output feature maps, respectively. This increase is due to the full PAON mapping that consists of \((P+Q)\) parallel convolutional branches, whereas a conventional convolution employs only a single branch. Figure~\ref{fig2} offers a comparative illustration of the computations involved in a convolutional, generative, and Padé neuron.

To enable end-to-end training of networks containing Padé neurons (PAONs), we derive the gradients of the loss function $\mathcal{L}$ with respect to the numerator kernels $\mathbf{w}_{pm,ik}^{(l)}$, the denominator kernels $\mathbf{w}_{qn,ik}^{(l)}$, and the output feature map of the previous layer $\mathbf{y}_i^{(l-1)}$. We denote the upstream gradient from the $(l+1)^{\mathrm{th}}$ layer as follows;
\begin{equation}
\boldsymbol{\delta}_k^{(l)} \triangleq 
\frac{\partial \mathcal{L}}{\partial \mathbf{x}_k^{(l)}},
\label{eq:delta_def}
\end{equation}
For compactness, we define:
\begin{align}
\mathbf{R}_k^{(l)} &= w_{p0,k}^{(l)}
+ \sum_{m=1}^{P}\sum_{i=1}^{N_{l-1}} 
\mathbf{w}_{pm,ik}^{(l)} \ast 
\big(\mathbf{y}_i^{(l-1)}\big)^{m}, \label{eq:bp_R}\\
\mathbf{S}_k^{(l)} &= 1 + \sum_{n=1}^{Q}\sum_{i=1}^{N_{l-1}}
\big|\,\mathbf{w}_{qn,ik}^{(l)} \ast 
\big(\mathbf{y}_i^{(l-1)}\big)^{n}\,\big|, \label{eq:bp_S}
\end{align}
so that $\mathbf{x}_k^{(l)} = \mathbf{R}_k^{(l)} \oslash \mathbf{S}_k^{(l)}$, where $\oslash$ denotes 
element-wise division and $\ast$ denotes 1D convolution. For any kernel $\mathbf{w}$, let 
$\tilde{\mathbf{w}}$ denote its time-reversal, $\tilde{\mathbf{w}}[t] = \mathbf{w}[-t]$. We use $\odot$ 
for element-wise multiplication, $[\cdot]^{-1}$ for element-wise inversion, $[\cdot]^{\circ 2}$ for 
element-wise squaring, $\ast_{\mathrm{grad}}$ for cross-correlation used in kernel gradients, and 
$\ast_{\mathrm{inp}}$ for convolution with $\tilde{\mathbf{w}}$ when backpropagating to inputs.

Since $\mathbf{w}_{pm,ik}^{(l)}$ appears only in $\mathbf{R}_k^{(l)}$, differentiating 
$\mathbf{x}_k^{(l)} = \mathbf{R}_k^{(l)} / \mathbf{S}_k^{(l)}$ with respect to $\mathbf{w}_{pm,ik}^{(l)}$ 
while treating $\mathbf{S}_k^{(l)}$ as constant yields;
\begin{equation}
\frac{\partial \mathcal{L}}{\partial \mathbf{w}_{pm,ik}^{(l)}} =
\left(\boldsymbol{\delta}_k^{(l)} \odot [\mathbf{S}_k^{(l)}]^{-1}\right)
\ast_{\mathrm{grad}} \big(\mathbf{y}_i^{(l-1)}\big)^{m}.
\label{eq:bp_num}
\end{equation}

For denominator kernels, we introduce 
$\mathbf{h}_{qn,ik}^{(l)} \triangleq \mathbf{w}_{qn,ik}^{(l)} \ast 
\big(\mathbf{y}_i^{(l-1)}\big)^{n}$ and use the subgradient 
$\operatorname{sgn}(\cdot)$ element-wise. 
Applying the quotient rule gives:
\begin{align}
\frac{\partial \mathcal{L}}{\partial \mathbf{w}_{qn,ik}^{(l)}} &=
\left(\boldsymbol{\delta}_k^{(l)} \odot
\big(-\,\mathbf{R}_k^{(l)} \oslash [\mathbf{S}_k^{(l)}]^{\circ 2}\big)\right)
\nonumber\\
&\quad \ast_{\mathrm{grad}}
\left(\operatorname{sgn}(\mathbf{h}_{qn,ik}^{(l)}) \odot
\big(\mathbf{y}_i^{(l-1)}\big)^{n}\right).
\label{eq:bp_den}
\end{align}

For the input gradients, both $\mathbf{R}_k^{(l)}$ and $\mathbf{S}_k^{(l)}$ 
depend on $\mathbf{y}_i^{(l-1)}$, leading to:
\begin{align}
\frac{\partial \mathbf{R}_k^{(l)}}{\partial \mathbf{y}_i^{(l-1)}} 
&= \sum_{m=1}^{P} 
\tilde{\mathbf{w}}_{pm,ik}^{(l)} \ast
\big( m\,(\mathbf{y}_i^{(l-1)})^{m-1} \big),
\label{eq:bp_R_inp}\\[2pt]
\frac{\partial \mathbf{S}_k^{(l)}}{\partial \mathbf{y}_i^{(l-1)}} 
&= \sum_{n=1}^{Q} 
\operatorname{sgn}\!\big(\mathbf{h}_{qn,ik}^{(l)}\big) \odot \nonumber\\
&\quad \left[\,\tilde{\mathbf{w}}_{qn,ik}^{(l)} \ast
\big( n\,(\mathbf{y}_i^{(l-1)})^{n-1} \big) \right].
\label{eq:bp_S_inp}
\end{align}

Finally, the gradient of the loss function with respect to the $i$-th output feature map $\mathbf{y}_i^{(l-1)}$ of the $(l-1)^{\mathrm{th}}$ layer can be expressed as:
\begin{align}
\frac{\partial \mathcal{L}}{\partial \mathbf{y}_i^{(l-1)}} 
&= \sum_{k=1}^{N_l} \boldsymbol{\delta}_k^{(l)} \odot \Big(
[\mathbf{S}_k^{(l)}]^{-1} \odot
\frac{\partial \mathbf{R}_k^{(l)}}{\partial \mathbf{y}_i^{(l-1)}} \nonumber\\
&\quad - (\mathbf{R}_k^{(l)} \oslash [\mathbf{S}_k^{(l)}]^{\circ 2}) \odot
\frac{\partial \mathbf{S}_k^{(l)}}{\partial \mathbf{y}_i^{(l-1)}} \Big).
\label{eq:bp_y_quotient}
\end{align}

\section{Experimental Evaluation} \label{sec3}
This section outlines the experimental setup used to acquire the University of Ottawa electric motor vibration and acoustic fault signature dataset (UOEMD-VAFCVS) \citep{sehri2024university}, describes the preprocessing steps applied to the data, and details the training and testing configurations, including data partitioning, training strategy, and evaluation criteria.

\begin{figure*}[htbp]
\includegraphics[width=\textwidth]{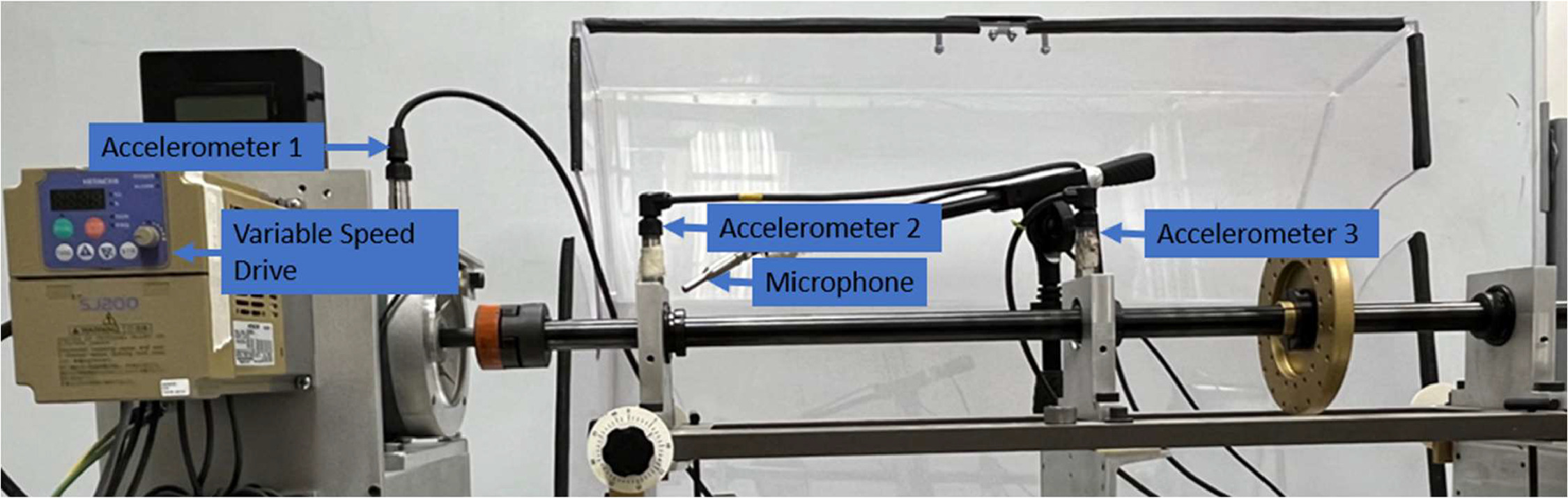}
\caption{The experimental setup \citep{sehri2024university}.
\label{fig3}}
\end{figure*}   

\subsection{Experimental Setup and Dataset} \label{sec3_1}
The University of Ottawa electric motor vibration and audio datasets were collected from a modified SpectraQuest Machinery Fault \& Rotor Dynamics Simulator test rig \citep{sehri2024university}. The setup includes an induction motor, a variable frequency drive, three single-axis accelerometers, and a microphone, as depicted in Figure~\ref{fig3}. The data collection system involves a National Instruments USB-6212 data acquisition unit to connect the sensors to a computer. The accelerometers measure vibration and temperature signals at both the drive end and shaft of the system, while the microphone captures acoustic signals. The variable frequency drive records the rotational speed of the motor. The data collection duration was fixed at 10 seconds, with data acquisition carried out using LabVIEW. All signals were sampled at a rate of 42 kHz \citep{sehri2024university}.

Each data file consists of time-series measurements organized into several columns. The first column contains data from the accelerometer (PCB, model 603C01) positioned at the drive end of the motor. The second column records acoustic data, while the third column presents data from a second accelerometer (PCB, model 623C01) positioned on the shaft's bearing housing near the drive end. The fourth column contains data from the third accelerometer (PCB, model 623C01), located on the shaft's bearing housing furthest from the drive end, and the remaining columns include temperature and rotational speed measurements \citep{sehri2024university}.

The dataset includes samples from a healthy induction motor and various fault conditions. These faults can be categorized as follows:
\begin{itemize}
    \item \textbf{Electrical faults:} stator winding faults (SW), voltage unbalance and single phasing (VU), and broken rotor bars (KA).
    \item \textbf{Mechanical faults:} rotor unbalance (RU), rotor misalignment (RM), bowed rotor (BR), and faulty bearings (FB).
\end{itemize}

Therefore, it contains a total of 8 classes, corresponding to the 7 fault conditions and the healthy motor. Induction machines were operated under both constant and variable operating frequencies. The constant frequencies were approximately 15 Hz, 30 Hz, 45 Hz, and 60 Hz. The variable frequencies included ranges such as 15 Hz to 45 Hz, 30 Hz to 60 Hz, 45 Hz to 15 Hz, and 60 Hz to 30 Hz. In this study, only the constant speed portion of the Ottawa dataset is utilized, and the subsequent discussions are based on this subset of the data. The motors were operated under both no-load and loaded conditions. The loading was implemented by symmetrically attaching ten bolts to a disk mounted on the motor shaft \citep{sehri2024university}. The dataset filenames follow the format {Letter}-{Letter}-{Number}-{Number}, where the first two letters indicate the motor’s condition (e.g., “H” for healthy, “R” for rotor fault, “B” for bowed rotor, etc.). The first number shows the motor speed setting (e.g., 1 = 15 Hz, 4 = 60 Hz), and the second number indicates the load condition (“0” for no load, “1” for loaded). For instance, “R-U-1-0” refers to an unloaded rotor unbalance fault at 15 Hz. The constant-speed dataset under both unloaded and loaded conditions can be structured as shown in Table~\ref{tab_dataset}.

\begin{table*}[htbp]
\small 
\caption{Labeling of datasets for constant speed conditions across different fault modes. The last digit indicates the load condition (0 = Unloaded, 1 = Loaded).}
\label{tab_dataset}
\begin{tabularx}{\linewidth}{l *{4}{>{\raggedright\arraybackslash}X}} 
\toprule
\textbf{Fault Mode} & \multicolumn{4}{c}{\textbf{Speed (Hz)}} \\
\cmidrule(r){2-5}
 & \textbf{15 Hz} & \textbf{30 Hz} & \textbf{45 Hz} & \textbf{60 Hz} \\
\midrule
Healthy (H) & \makecell[l]{H-H-1-0 \\ H-H-1-1} & \makecell[l]{H-H-2-0 \\ H-H-2-1} & \makecell[l]{H-H-3-0 \\ H-H-3-1} & \makecell[l]{H-H-4-0 \\ H-H-4-1} \\
\midrule
Rotor Unbalance (RU) & \makecell[l]{R-U-1-0 \\ R-U-1-1} & \makecell[l]{R-U-2-0 \\ R-U-2-1} & \makecell[l]{R-U-3-0 \\ R-U-3-1} & \makecell[l]{R-U-4-0 \\ R-U-4-1} \\
\midrule
Rotor Misalignment (RM) & \makecell[l]{R-M-1-0 \\ R-M-1-1} & \makecell[l]{R-M-2-0 \\ R-M-2-1} & \makecell[l]{R-M-3-0 \\ R-M-3-1} & \makecell[l]{R-M-4-0 \\ R-M-4-1} \\
\midrule
Stator Winding Fault (SW) & \makecell[l]{S-W-1-0 \\ S-W-1-1} & \makecell[l]{S-W-2-0 \\ S-W-2-1} & \makecell[l]{S-W-3-0 \\ S-W-3-1} & \makecell[l]{S-W-4-0 \\ S-W-4-1} \\
\midrule
Voltage Unbalance (VU) & \makecell[l]{V-U-1-0 \\ V-U-1-1} & \makecell[l]{V-U-2-0 \\ V-U-2-1} & \makecell[l]{V-U-3-0 \\ V-U-3-1} & \makecell[l]{V-U-4-0 \\ V-U-4-1} \\
\midrule
Bowed Rotor (BR) & \makecell[l]{B-R-1-0 \\ B-R-1-1} & \makecell[l]{B-R-2-0 \\ B-R-2-1} & \makecell[l]{B-R-3-0 \\ B-R-3-1} & \makecell[l]{B-R-4-0 \\ B-R-4-1} \\
\midrule
Broken Rotor Bars (KA) & \makecell[l]{K-A-1-0 \\ K-A-1-1} & \makecell[l]{K-A-2-0 \\ K-A-2-1} & \makecell[l]{K-A-3-0 \\ K-A-3-1} & \makecell[l]{K-A-4-0 \\ K-A-4-1} \\
\midrule
Faulty Bearings (FB) & \makecell[l]{F-B-1-0 \\ F-B-1-1} & \makecell[l]{F-B-2-0 \\ F-B-2-1} & \makecell[l]{F-B-3-0 \\ F-B-3-1} & \makecell[l]{F-B-4-0 \\ F-B-4-1} \\
\bottomrule
\end{tabularx}
\end{table*}

\subsection{Data Preparation} \label{sec3_2}

In this work, we evaluate the performance of 1D PadéNets for vibration and audio inputs separately. For each input channel, the entire dataset was partitioned into training (80\%), validation (10\%), and testing (10\%) subsets through a sequential temporal split performed separately on each individual signal file. Each file, corresponding to a distinct operating frequency and fault class, was split temporally such that the initial 80\% of its segments were used for training, the subsequent 10\% for validation, and the remaining 10\% for testing. This approach ensures that the model is evaluated on future unseen data and avoids temporal leakage, which can occur if segments from the same temporal window appear in both training and evaluation sets. No shuffling was performed either within or between files. Finally, the segments were categorized into fault classes, with each class containing data collected across all operating frequencies and both loaded and unloaded machine conditions. The number of samples in each split is summarized in Table~\ref{tab_numsamples}.

\begin{table*}[htbp] 
\caption{Number of samples in each data split for each input channel, including both unloaded and loaded conditions of the constant-speed datasets. \label{tab_numsamples}}
\begin{tabularx}{\linewidth}{cccc}
\toprule
\textbf{Fault Mode}	& \textbf{Training Samples}	& \textbf{Validation Samples} & \textbf{Test Samples}\\
\midrule
Healthy (H)		& 2680			& 336    & 336\\
Rotor Unbalance (RU)		& 2680			& 336    & 336\\
Rotor Misalignment (RM)		& 2680			& 336    & 336\\
Stator Winding Fault (SW)		& 2680			& 336    & 336\\
Voltage Unbalance (VU)		& 2680			& 336    & 336\\
Bowed Rotor (BR)		& 2680			& 336    & 336\\
Broken Rotor Bars (KA)		& 2680			& 336    & 336\\
Faulty Bearings (FB)		& 2680			& 336    & 336\\
\bottomrule
\end{tabularx}
\end{table*}

\begin{figure*}[htbp]
\centering
\subfloat[\centering]{\includegraphics[width=0.5\linewidth]{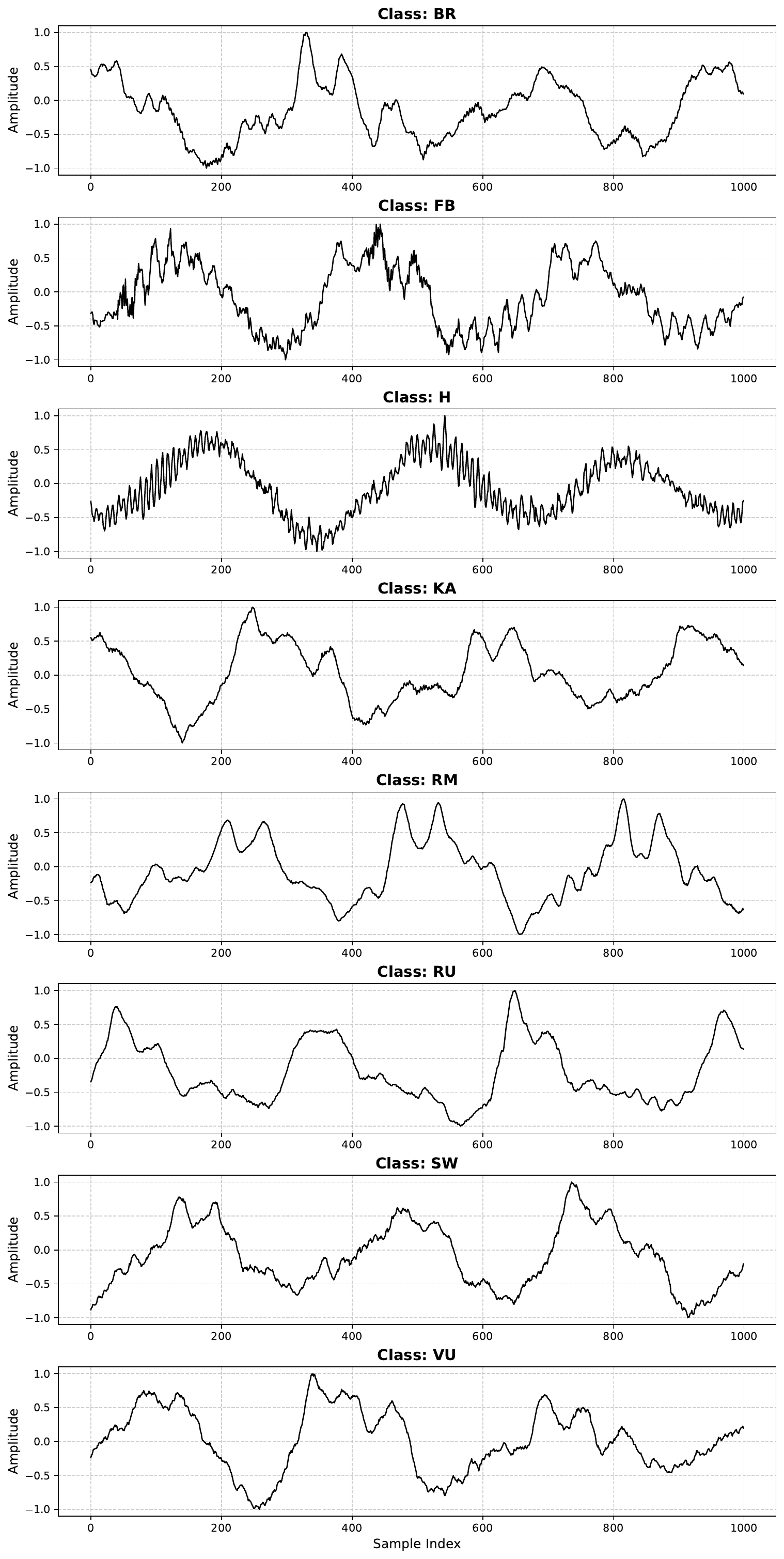}}
\subfloat[\centering]{\includegraphics[width=0.5\linewidth]{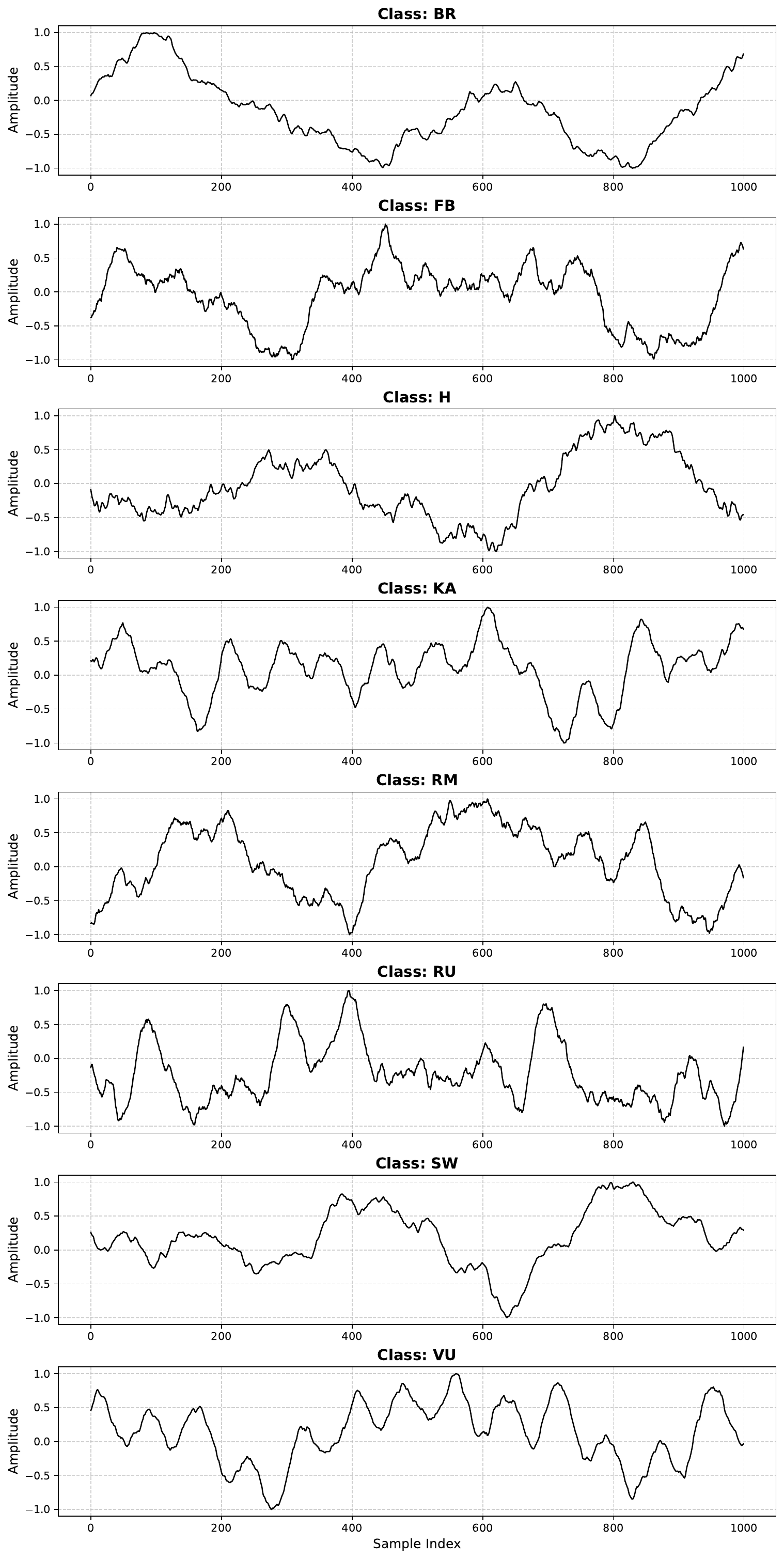}}\\
\caption{Examples of segmented and normalized waveforms for 45 Hz input frequency: (\textbf{a}) Accelerometer-1, (\textbf{b}) microphone. \label{fig4}}
\end{figure*}

All channels of vibration and audio data were segmented into fixed-length windows of 1000 time-domain samples without overlap. Thus, a total of 420 samples were extracted per channel from a single recording. The segmentation process divides the continuous time series data into smaller, non-overlapping segments of 1000 samples to balance temporal resolution and model complexity. Each segment is then normalized separately. Normalization is performed using the min-max normalization method, where each data point \(x_i\) in a given segment is scaled based on the minimum and maximum values of that segment. The normalization equation can be written as:
\begin{equation}
x_i^{\text{norm}} = \dfrac{2 \cdot (x_i - x_{\text{min}})}{x_{\text{max}} - x_{\text{min}}} - 1, 
\label{eq10}
\end{equation}
where \(x_i^{\text{norm}}\) is the normalized value, \(x_i\) is the original data point, \(x_{\text{min}}\) and \(x_{\text{max}}\) are the minimum and maximum values of the segment, respectively. Figure~\ref{fig4} presents some examples of segmented and normalized vibration and audio waveforms obtained from the first accelerometer and microphone at an operating frequency of 45 Hz.

\subsection{Model Architecture and Training Strategy} \label{sec3_3}
The architecture of the proposed 1D PadéNet-based framework is illustrated in Figure~\ref{fig1} and detailed in Table~\ref{tab_padenet}. In this context, \( P \) and \( Q \) are hyperparameters that define the degree of the nonlinear transformations applied within the 1D Padé layers. Specifically, \( P \) determines the order of the numerator polynomial, and \( Q \) determines the order of the denominator polynomial in the Padé approximation used in these layers. By adjusting \( P \) and \( Q \), the model can be tailored to capture different levels of nonlinearity in the data, allowing for enhanced feature extraction. For a direct comparison with 1D CNNs, we set \( P = 1 \) and \( Q = 0 \), which simplifies the Padé neuron to a standard convolutional neuron. On the other hand, for comparison with 1D Self-ONNs, we set \( P \geq 2 \) and \( Q = 0 \), allowing the Padé neurons to function as generative neurons, as used in Self-ONNs.

\begin{table*}[htbp] 
\caption{Detailed architecture of the proposed 1D PadéNet model is given. Activations in the 1D Padé layers ($\tanh$ and $\mathrm{LeakyReLU}$ with negative slope coefficient 0.01) were evaluated separately and not explicitly shown. \label{tab_padenet}}
\begin{tabularx}{\linewidth}{ccccc}
\toprule
\textbf{Number} & \textbf{Layer Name} & \textbf{Filters / Kernel Size} & \textbf{Output Shape} & \textbf{Other Hyperparams.} \\
\midrule
1 & 1D Padé & 32 / 7 & (None, 1000, 32) & $P, Q$, Strides = 1 \\
2 & 1D MaxPooling & – & (None, 500, 32) & Pool Size = 2 \\
3 & 1D Padé & 32 / 7 & (None, 500, 32) & $P, Q$, Strides = 1 \\
4 & 1D MaxPooling & – & (None, 250, 32) & Pool Size = 2 \\
5 & 1D Padé & 32 / 7 & (None, 250, 32) & $P, Q$, Strides = 1 \\
6 & 1D MaxPooling & – & (None, 125, 32) & Pool Size = 2 \\
7 & 1D Padé & 32 / 7 & (None, 125, 32) & $P, Q$, Strides = 1 \\
8 & 1D MaxPooling & – & (None, 62, 32) & Pool Size = 2 \\
9 & 1D Padé & 32 / 7 & (None, 62, 32) & $P, Q$, Strides = 1 \\
10 & 1D MaxPooling & – & (None, 31, 32) & Pool Size = 2 \\
11 & 1D Padé & 32 / 7 & (None, 31, 32) & $P, Q$, Strides = 1 \\
12 & 1D MaxPooling & – & (None, 15, 32) & Pool Size = 2 \\
13 & 1D Padé & 32 / 7 & (None, 15, 32) & $P, Q$, Strides = 1 \\
14 & 1D MaxPooling & – & (None, 7, 32) & Pool Size = 2 \\
15 & Flatten & – & (None, 224) & – \\
16 & Dropout & – & (None, 224) & Dropout Rate = 0.25 \\
17 & Dense & – & (None, 64) & Activation = $\tanh$ \\
18 & Dense & – & (None, 8) & Activation = Softmax \\
\bottomrule
\end{tabularx}
\end{table*}

Following each 1D Padé layer, the activation function applied is either the $\tanh$ or $\mathrm{LeakyReLU}$ with a negative slope of 0.01. We evaluate the performance of both activation functions separately for 1D CNN and PadéNet. However, in 1D Self-ONNs, we are restricted to using $\tanh$ to prevent excessively large activations due to the increasing powers of the input, thus mitigating the risk of the exploding gradients. For all models, a 1D max-pooling layer with a pool size of 2 was applied after each block for spatial downsampling, reducing the dimensionality of the feature maps while preserving the most crucial information.

Each of the seven 1D Padé layers was configured with 32 filters, a kernel size of 7, padding='same', and incorporated an L2 kernel regularizer (\(\lambda = 10^{-4}\)) for weight decay. Following these layers, a Flatten layer was used to reshape the feature maps into a vector. It was followed by a dense layer with 64 neurons and a $\tanh$ activation, which processes the flattened features. The model ends with a Softmax output layer with 8 neurons for multi-class classification, assigning probabilities to each of the 8 classes. To prevent overfitting, a Dropout layer with a rate of 0.25 was applied before the dense layers during training.

Training was performed over a maximum of 100 epochs with a batch size of 64. The network parameters were optimized using the Adam optimizer with an initial learning rate of 0.0005 and the categorical cross-entropy loss function. Early stopping was employed to stop training if the validation loss did not improve over 20 consecutive epochs. Additionally, a learning rate decay callback was used to reduce the learning rate by a factor of 0.5 if there was no reduction in validation loss over 10 epochs. To evaluate each model’s classification accuracy and robustness, we conducted 5 independent runs with different random seeds. For each run, both the training and validation data were reshuffled using a fixed seed to ensure reproducibility.

\subsection{Evaluation Metrics} \label{sec3_4}

The fault diagnosis performance of each model was evaluated on the test set using the following classification metrics: Accuracy, Precision, Recall, and F1-Score. To account for variability due to random initialization, each model was trained and evaluated across 5 independent runs with distinct random seeds. For each metric, the mean and standard deviation were reported to summarize performance, along with the minimum and maximum accuracy values.

Accuracy represents the overall proportion of correctly predicted instances, calculated as the ratio of the sum of true positives (TP) and true negatives (TN) to the total number of samples as follows:
\begin{equation}
\text{Accuracy} = \frac{TP + TN}{TP + TN + FP + FN}
\label{eq11}
\end{equation}

Precision is also important when the cost of false positives (i.e., incorrectly identifying a motor fault when there is not one) is high. It is defined as the ratio of true positives to the total number of predicted positives (TP + FP). High precision ensures that the model does not produce too many false alarms, which is important in scenarios where an incorrect fault diagnosis could lead to unnecessary maintenance or costly downtime. On the other hand, Recall measures the model’s ability to correctly identify all actual positives (i.e., actual faults). It is defined as the ratio of true positives to the total number of actual positives (TP + FN). Recall is particularly important in fault diagnosis because failing to identify a fault (i.e., a false negative) could lead to catastrophic consequences, such as equipment failure or unplanned downtime. The equations for Precision and Recall are given as:
\begin{equation}
\text{Precision} = \frac{TP}{TP + FP}, \quad \text{Recall} = \frac{TP}{TP + FN}
\label{eq12}
\end{equation}

Finally, the F1-Score is the harmonic mean of Precision and Recall, providing a balanced evaluation of both metrics. It can be computed as:
\begin{equation}
\text{F1\text{-}Score} = 2 \cdot \frac{\text{Precision} \cdot \text{Recall}}{\text{Precision} + \text{Recall}}
\label{eq13}
\end{equation}

By using these metrics in combination, we can obtain a comprehensive evaluation of each model's performance to have a clear understanding of its effectiveness for motor fault diagnosis.

\section{Results and Discussion} \label{sec4}
This section presents the fault diagnosis performance of Padé-based models, along with CNNs and Self-ONNs, on the constant-speed vibration and acoustic datasets from Ottawa University \citep{sehri2024university}. The analysis further evaluates robustness under varying noise levels, examines the sensitivity of Padé layers to different Padé orders and activation functions, and compares the corresponding computational complexities.

\subsection{Performance on the Ottawa University Vibration and Acoustic Datasets} \label{sec4_1}

\begin{figure*}[htbp]
\centering
\subfloat[\centering]{\includegraphics[width=0.33\linewidth]{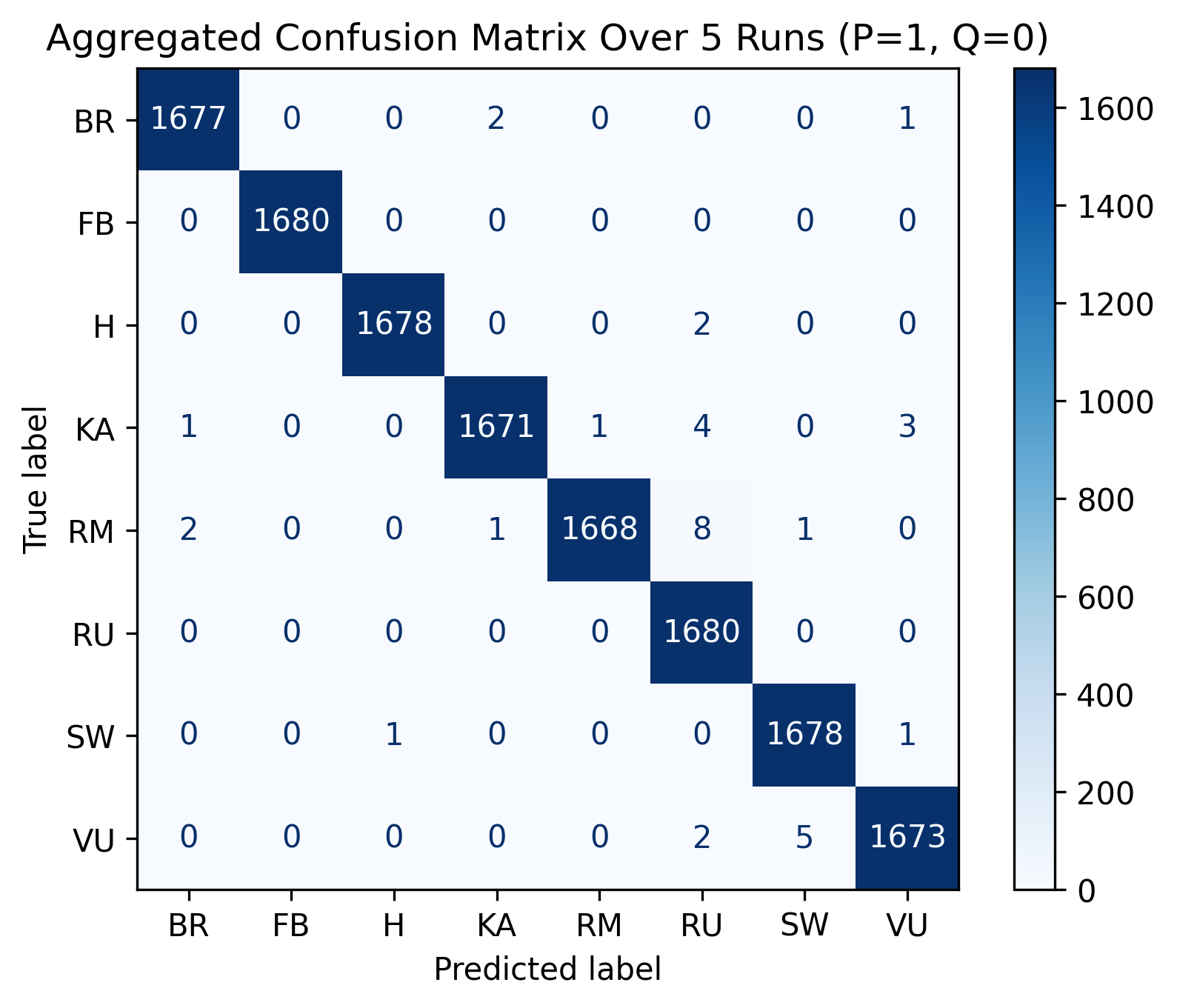}}
\hfill
\subfloat[\centering]{\includegraphics[width=0.33\linewidth]{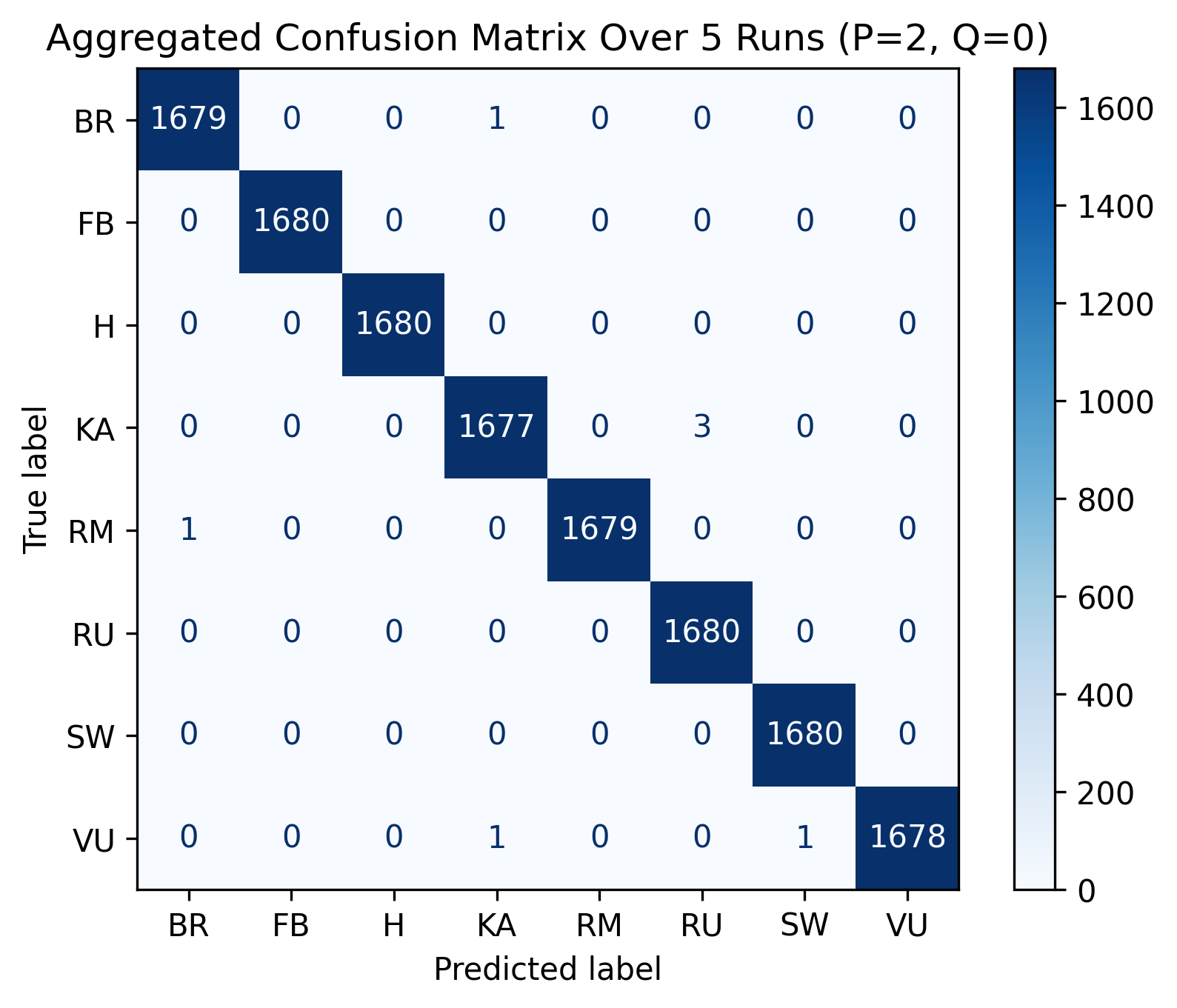}}
\hfill
\subfloat[\centering]{\includegraphics[width=0.33\linewidth]{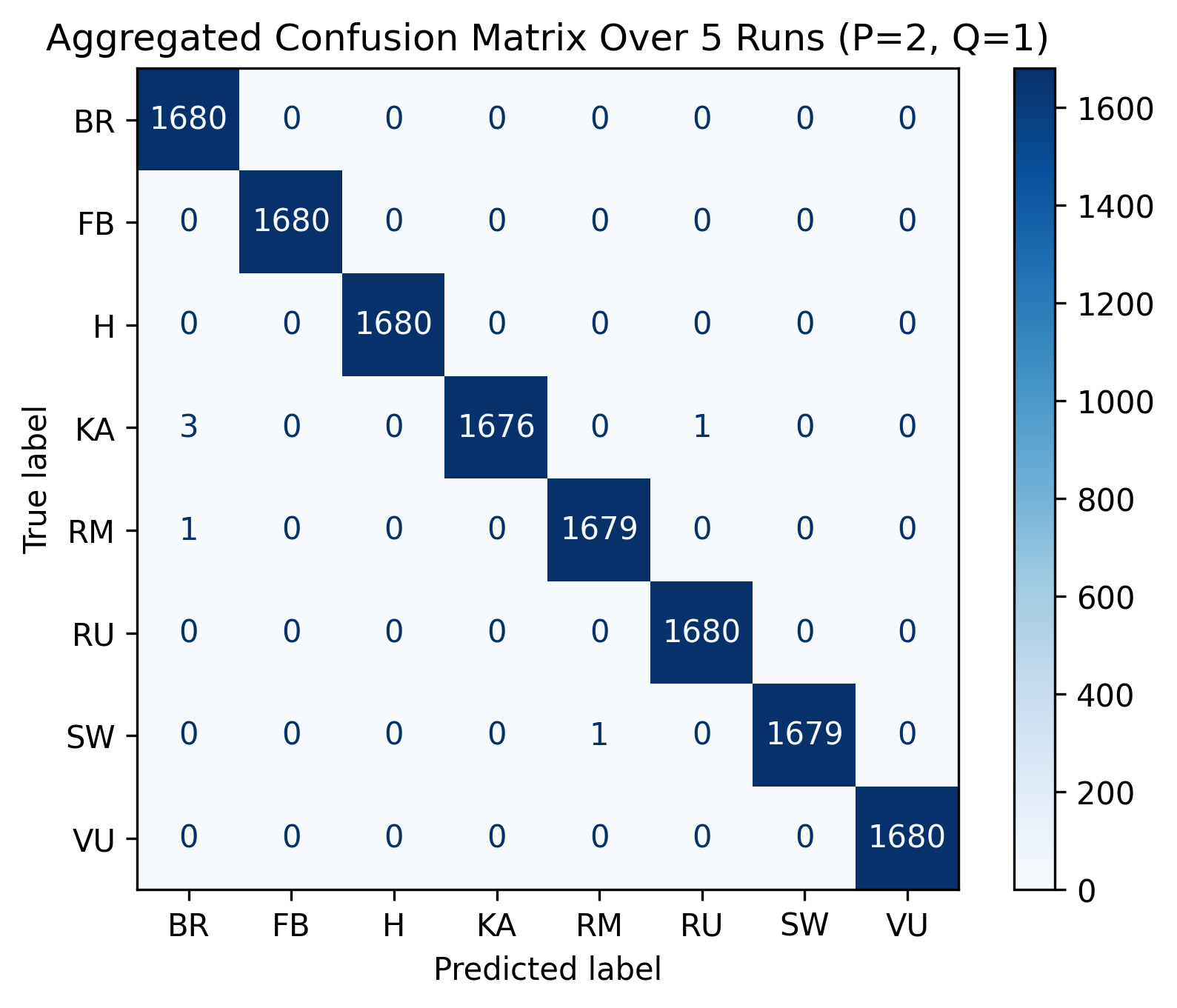}}
\caption{Aggregated confusion matrices over 5 runs for the 1D CNN ($P=1, Q=0$), Self-ONN ($P=2, Q=0$), and PadéNet ($P=2, Q=1$) models with the first accelerometer data as input. The 1D Self-ONN and PadéNet models use the $\tanh$ activation function, while the 1D CNN utilizes $\mathrm{LeakyReLU}$ with negative slope of 0.01.} 
\label{fig5}
\end{figure*}

\begin{figure*}[htbp]
\centering
\subfloat[\centering]{\includegraphics[width=0.33\linewidth]{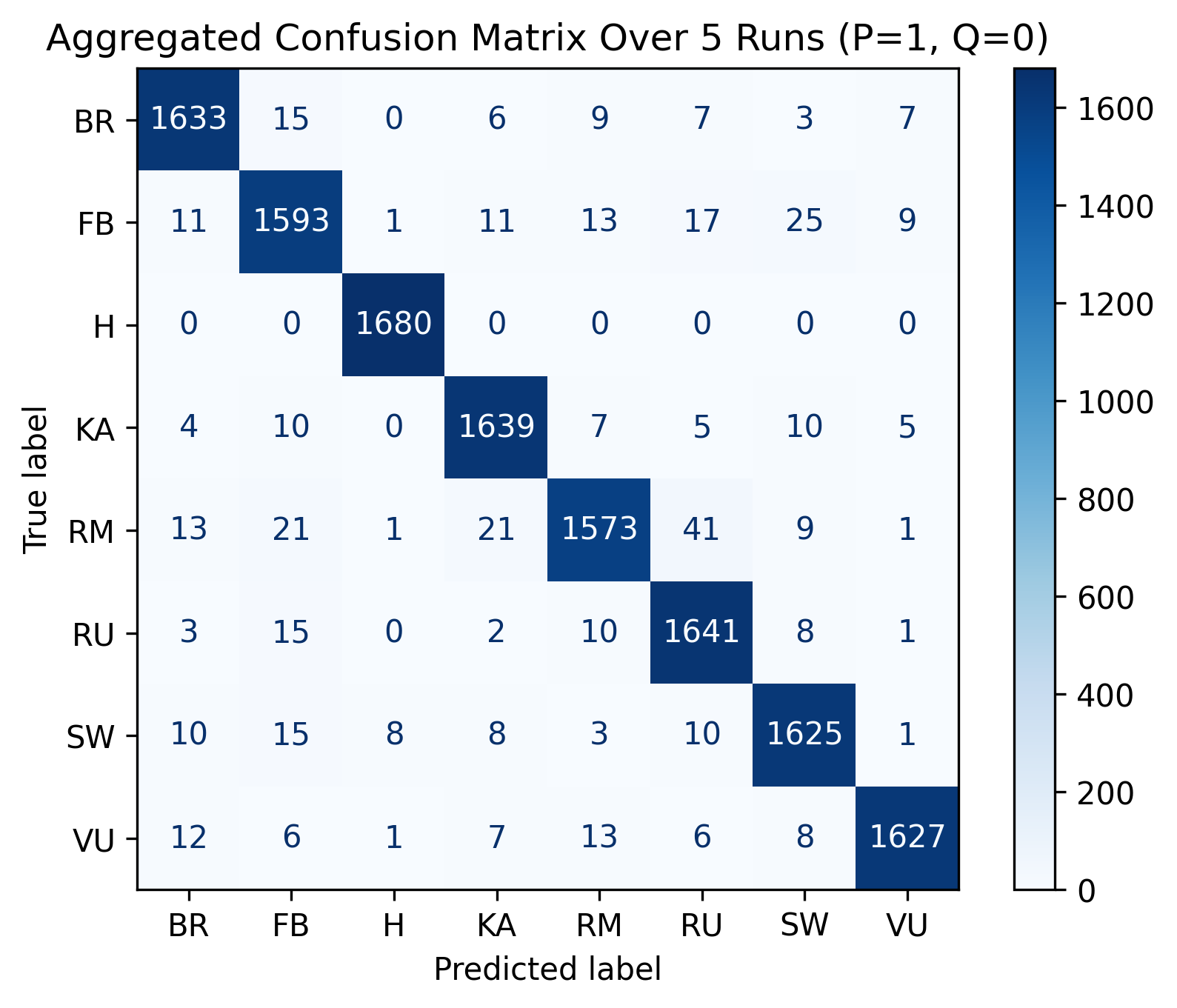}}
\hfill
\subfloat[\centering]{\includegraphics[width=0.33\linewidth]{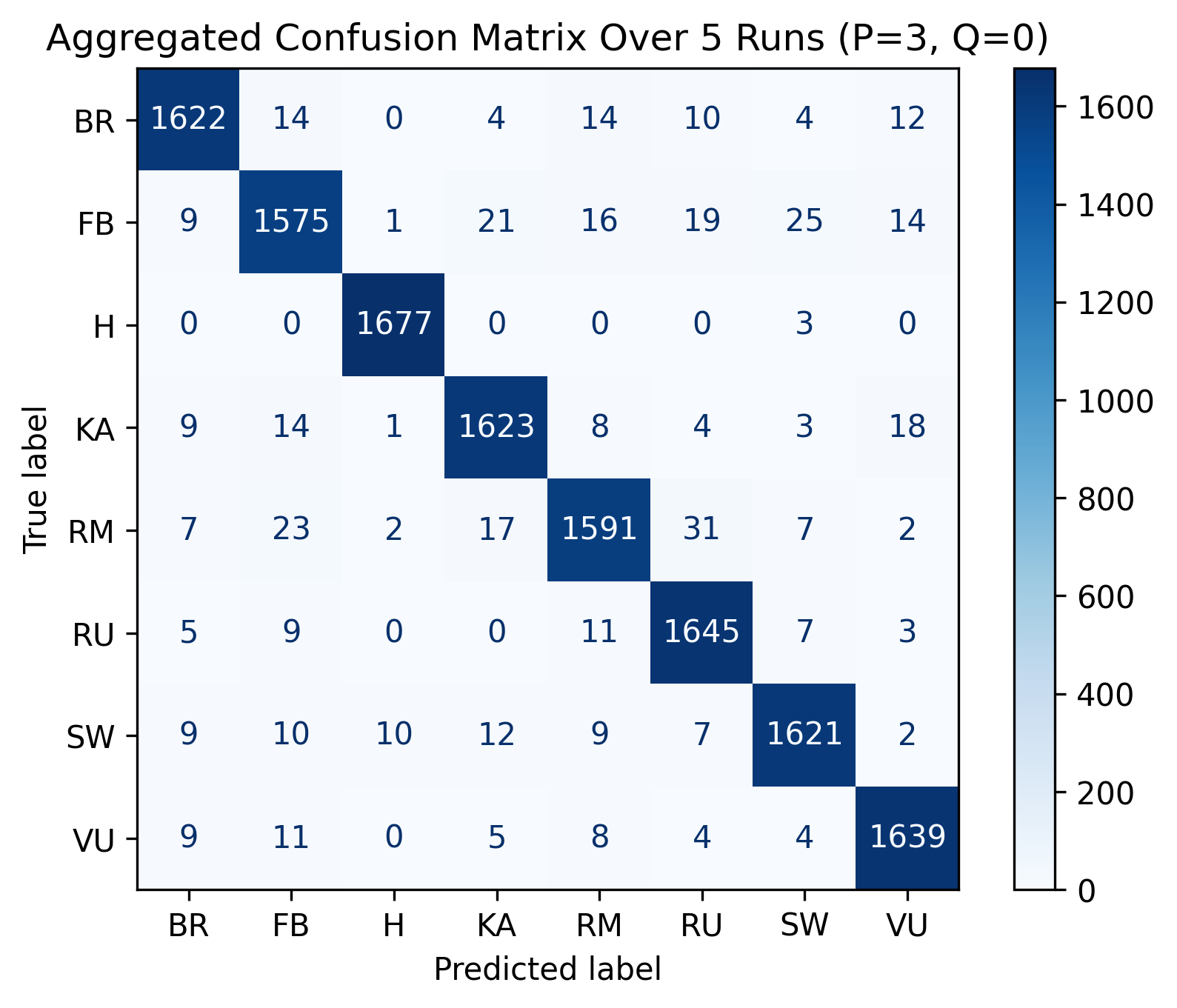}}
\hfill
\subfloat[\centering]{\includegraphics[width=0.33\linewidth]{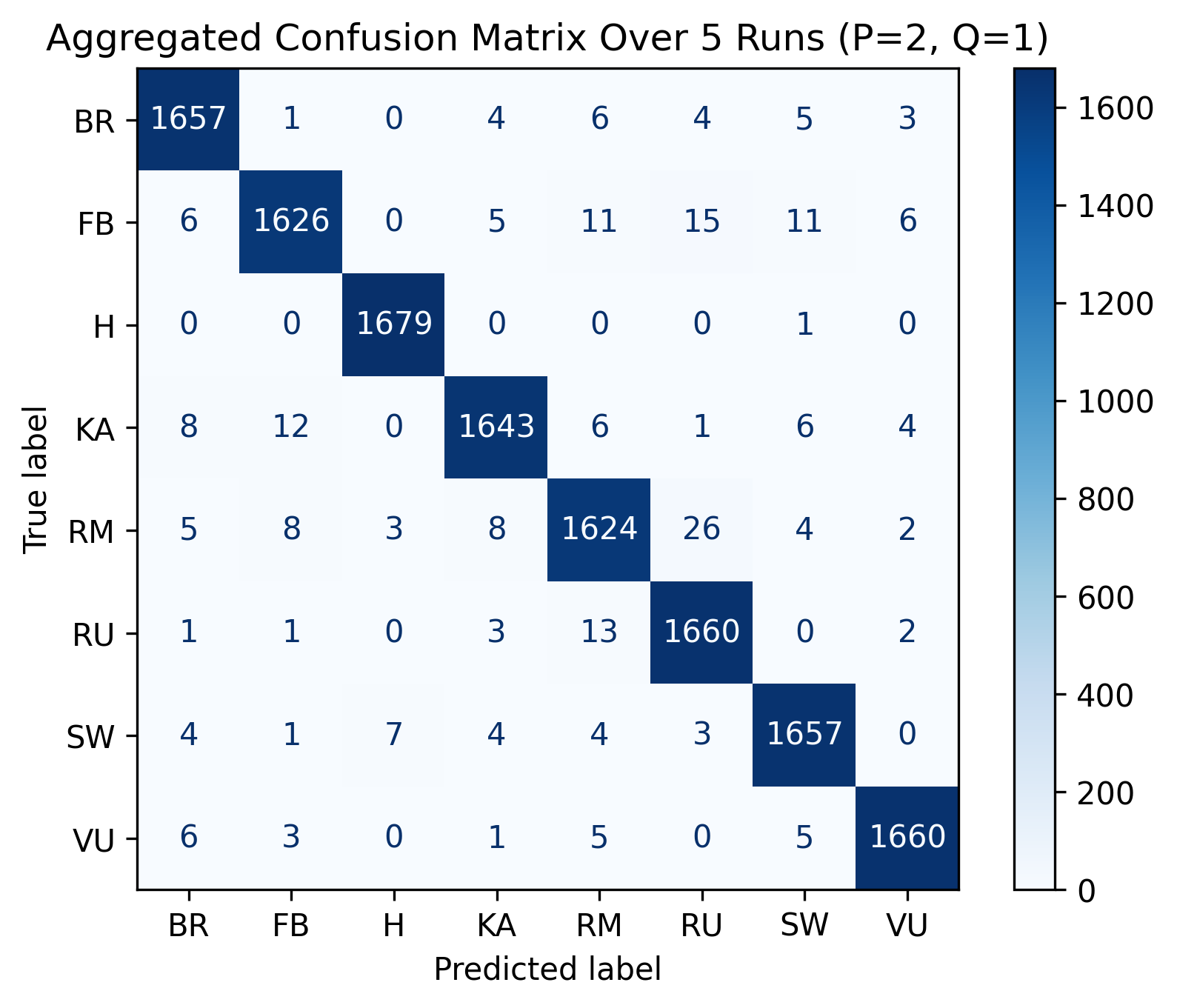}}
\caption{Aggregated confusion matrices over 5 runs for the 1D CNN ($P=1, Q=0$), Self-ONN ($P=3, Q=0$), and PadéNet ($P=2, Q=1$) models with the second accelerometer data as input. The 1D Self-ONN model uses the $\tanh$ activation function, while the 1D CNN and PadéNet utilize $\mathrm{LeakyReLU}$ with negative slope of 0.01.} 
\label{fig6}
\end{figure*}

\begin{figure*}[htbp]
\centering
\subfloat[\centering]{\includegraphics[width=0.33\linewidth]{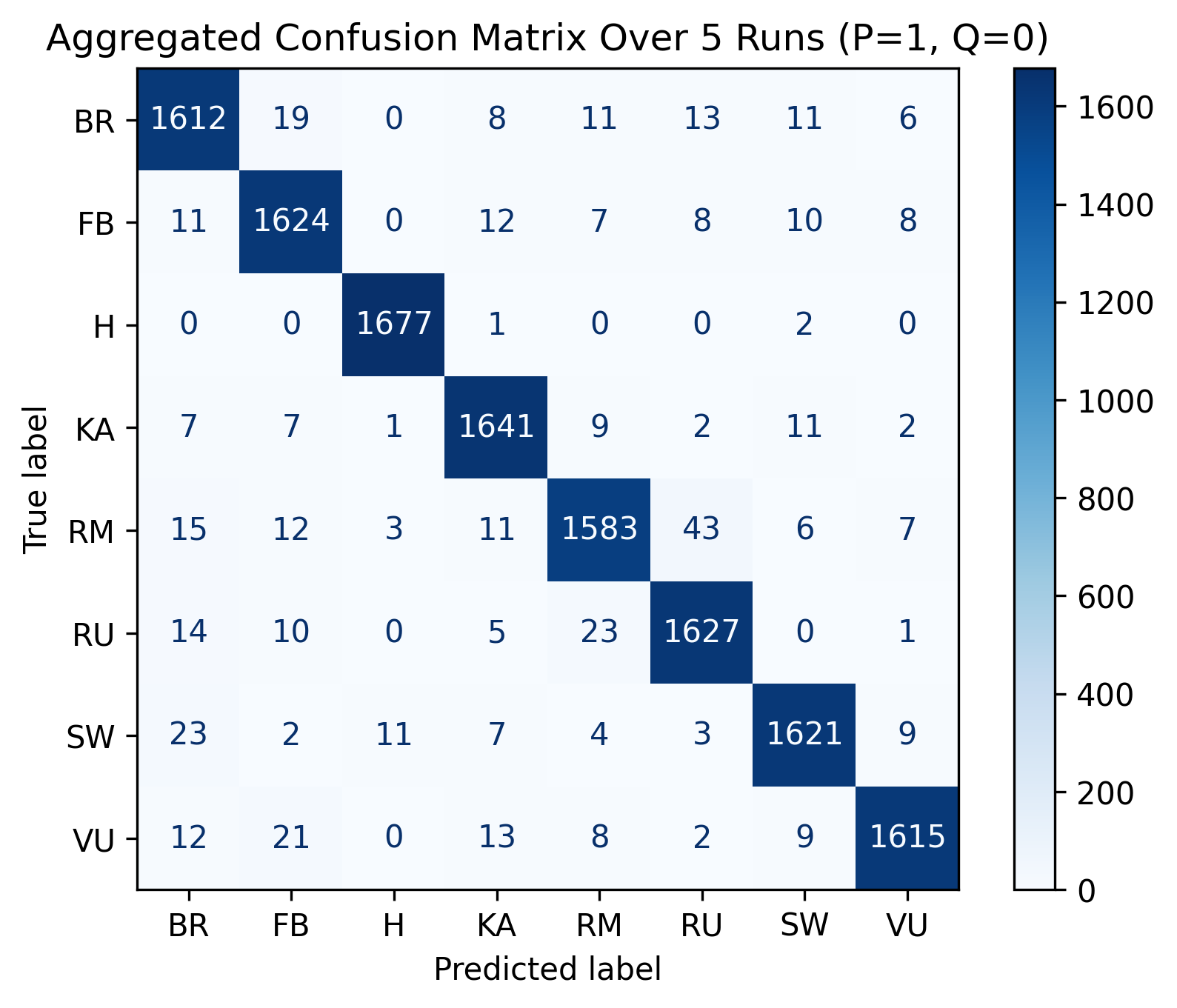}}
\hfill
\subfloat[\centering]{\includegraphics[width=0.33\linewidth]{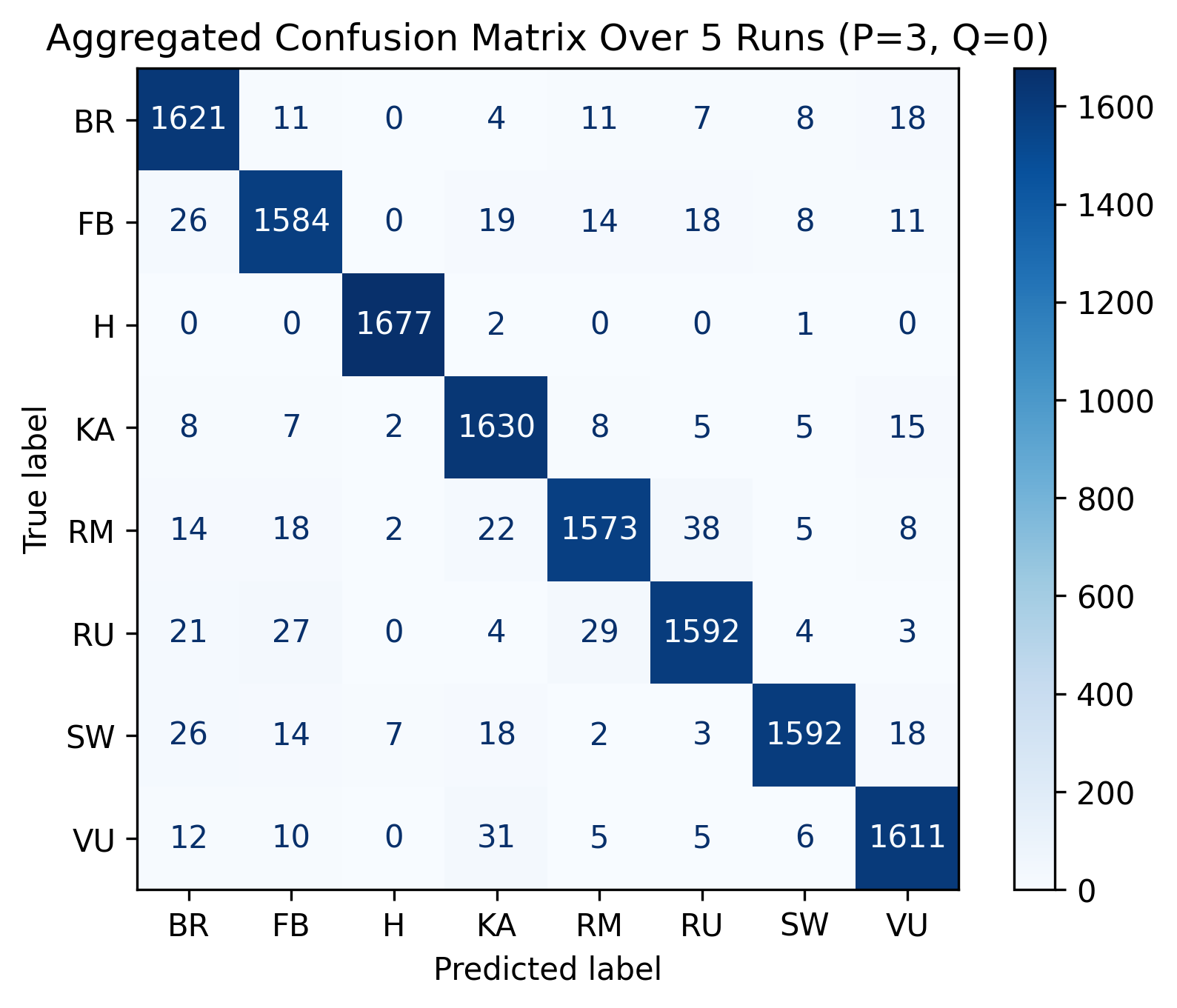}}
\hfill
\subfloat[\centering]{\includegraphics[width=0.33\linewidth]{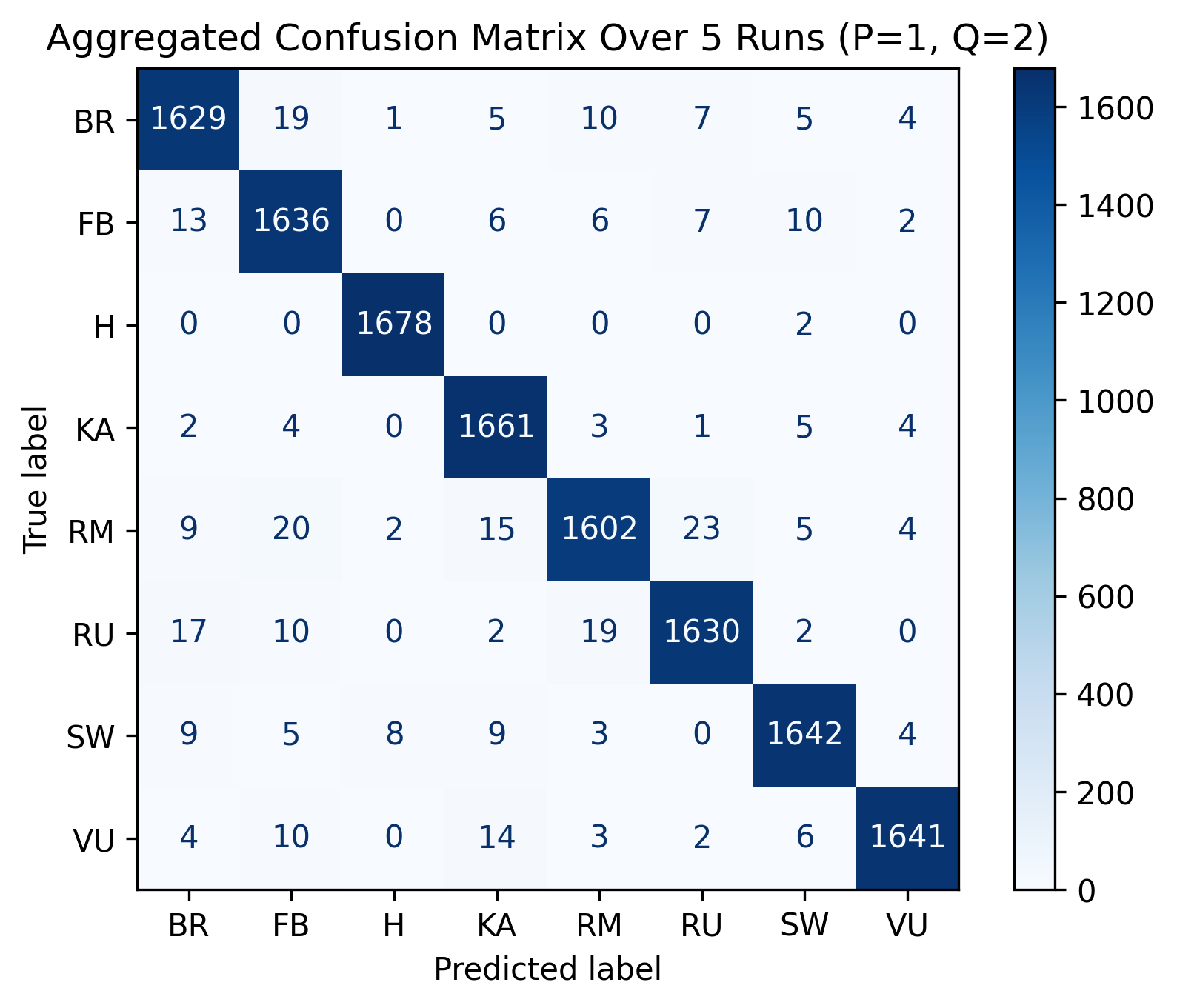}}
\caption{Aggregated confusion matrices over 5 runs for the 1D CNN ($P=1, Q=0$), Self-ONN ($P=3, Q=0$), and PadéNet ($P=1, Q=2$) models with the third accelerometer data as input. The 1D Self-ONN model uses the $\tanh$ activation function, while the 1D CNN and PadéNet utilize $\mathrm{LeakyReLU}$ with negative slope of 0.01.} 
\label{fig7}
\end{figure*}

\begin{figure*}[htbp]
\centering
\subfloat[\centering]{\includegraphics[width=0.33\linewidth]{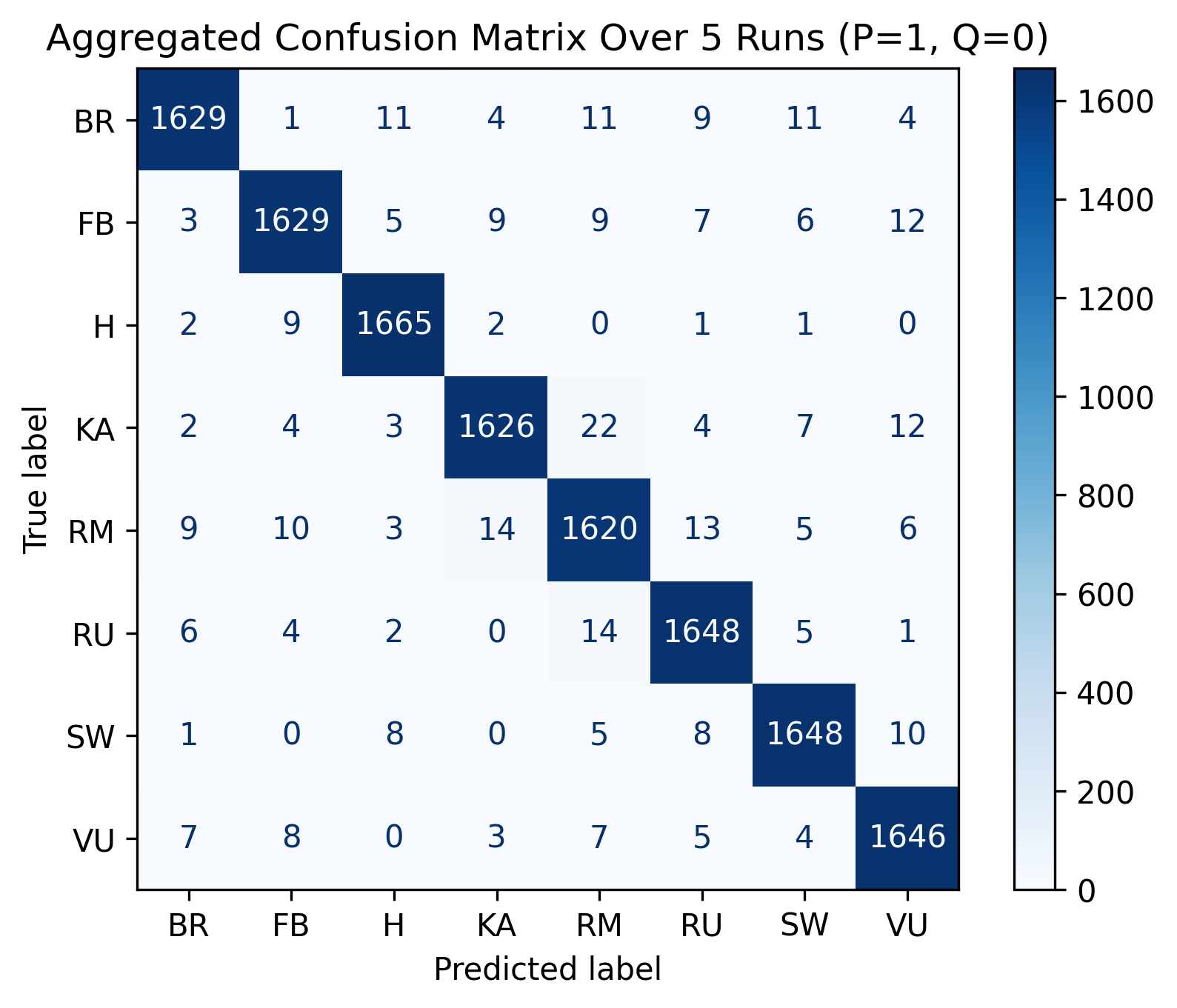}}
\hfill
\subfloat[\centering]{\includegraphics[width=0.33\linewidth]{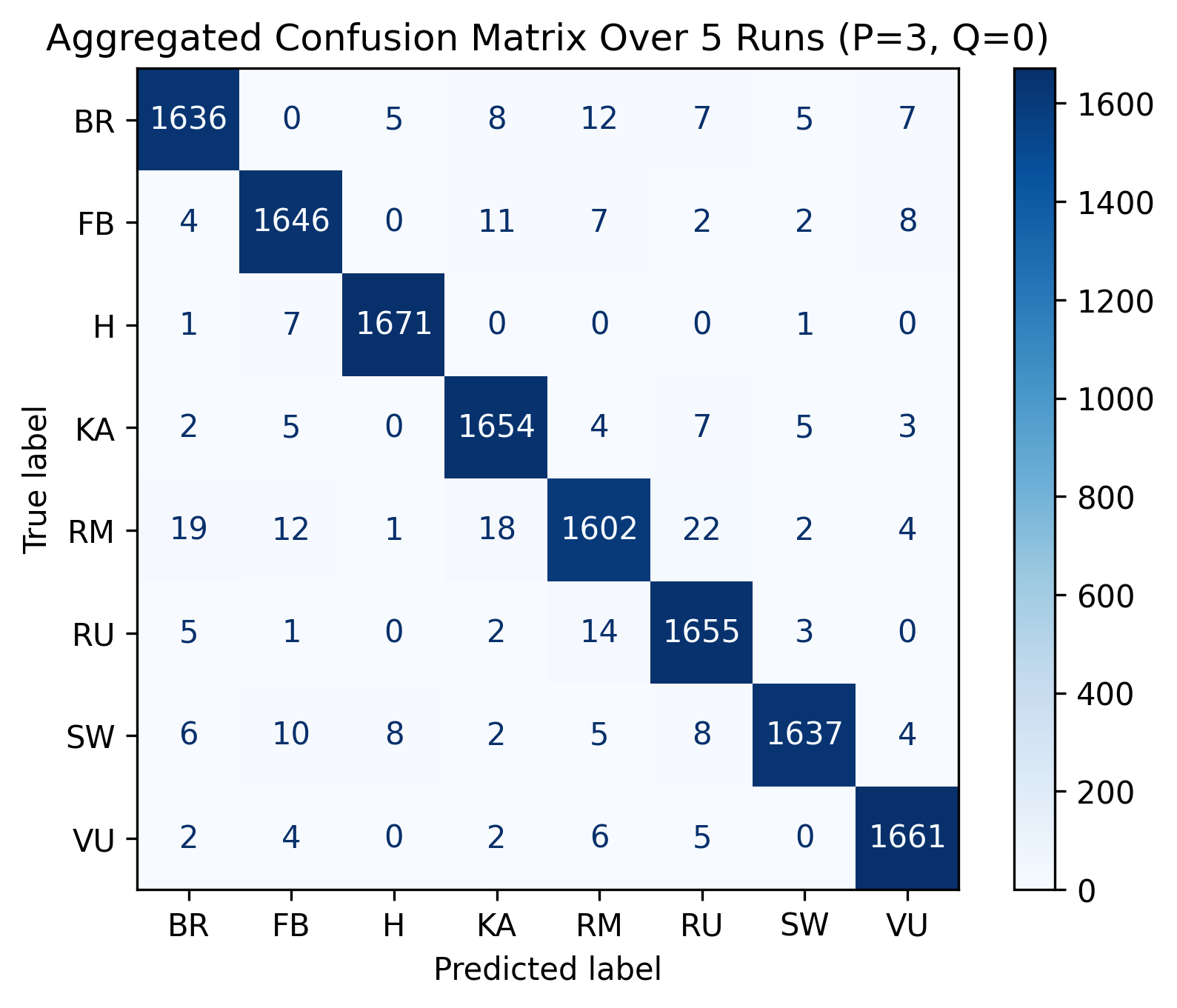}}
\hfill
\subfloat[\centering]{\includegraphics[width=0.33\linewidth]{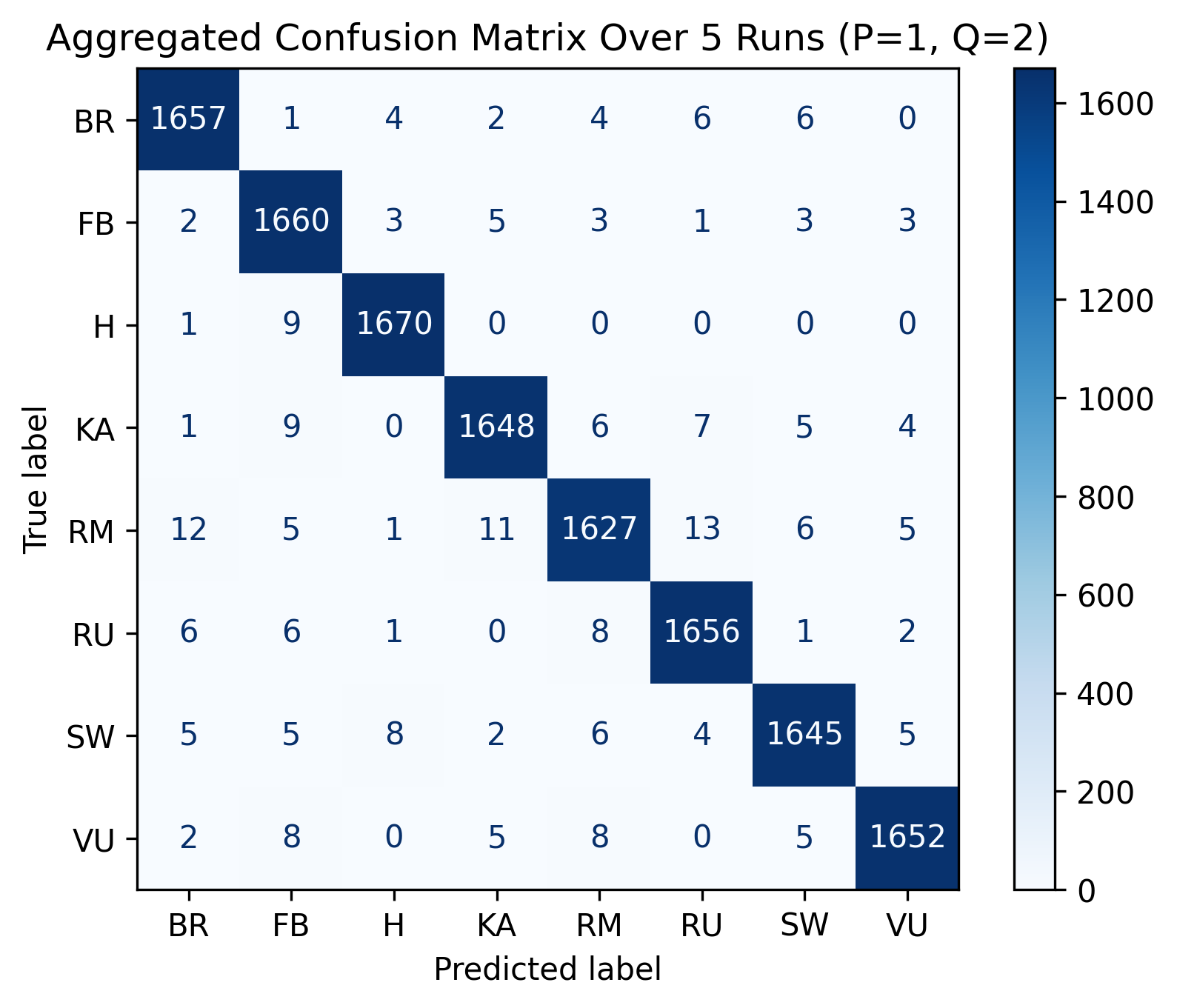}}
\caption{Aggregated confusion matrices over 5 runs for the 1D CNN ($P=1, Q=0$), Self-ONN ($P=3, Q=0$), and PadéNet ($P=1, Q=2$) models with the acoustic data as input. The 1D Self-ONN model uses the $\tanh$ activation function, while the 1D CNN and PadéNet utilize $\mathrm{LeakyReLU}$ with negative slope of 0.01.} 
\label{fig8}
\end{figure*}

We first present a comprehensive evaluation of the proposed 1D PadéNet compared to 1D CNN and Self-ONN using noise-free vibration and acoustic data \citep{sehri2024university}. Performance metrics were computed for each sensor input on 5 independent runs, and the results were averaged, with standard deviations reported to quantify variability, as detailed in Tables~\ref{tab_1staccel}, \ref{tab_2ndaccel}, \ref{tab_3rdaccel}, and \ref{tab_audio}. Aggregated confusion matrices over 5 runs (Figures~\ref{fig5}, \ref{fig6}, \ref{fig7}, \ref{fig8}) provide further insight into classification performance across fault types.

\begin{table*}[htbp]
\centering
\caption{Average test performance metrics (over 5 seeds) for different Padé models using the first accelerometer data as input.}
\label{tab_1staccel}
\renewcommand{\arraystretch}{1.1}
\setlength{\tabcolsep}{5pt}
\begin{tabularx}{\linewidth}{l *{6}{>{\centering\arraybackslash}X}}
\toprule
\textbf{Padé Model} & \textbf{Min. Acc. (\%)} & \textbf{Max. Acc. (\%)} & \textbf{Avg. Acc. (\%)} & \textbf{Avg. Prec. (\%)} & \textbf{Avg. Rec. (\%)} & \textbf{Avg. F1 (\%)} \\
\midrule
$P=1, Q=0$ (tanh)               & 99.03 & 100.00 & 99.69{\scriptsize$\pm$0.35} & 99.70{\scriptsize$\pm$0.33} & 99.69{\scriptsize$\pm$0.35} & 99.69{\scriptsize$\pm$0.35} \\
$P=1, Q=0$ (LeakyReLU)          & 99.33 & 100.00 & 99.74{\scriptsize$\pm$0.22} & 99.74{\scriptsize$\pm$0.22} & 99.74{\scriptsize$\pm$0.22} & 99.74{\scriptsize$\pm$0.22} \\
$P=1, Q=1$ (tanh)               & 99.85 & 100.00 & 99.95{\scriptsize$\pm$0.06} & 99.95{\scriptsize$\pm$0.06} & 99.95{\scriptsize$\pm$0.06} & 99.95{\scriptsize$\pm$0.06} \\
$P=1, Q=1$ (LeakyReLU)          & 99.89 &  99.96 & 99.92{\scriptsize$\pm$0.03} & 99.92{\scriptsize$\pm$0.03} & 99.92{\scriptsize$\pm$0.03} & 99.92{\scriptsize$\pm$0.03} \\
$P=1, Q=2$ (tanh)               & 99.85 & 100.00 & 99.96{\scriptsize$\pm$0.05} & 99.96{\scriptsize$\pm$0.05} & 99.96{\scriptsize$\pm$0.05} & 99.96{\scriptsize$\pm$0.05} \\
$P=1, Q=2$ (LeakyReLU)          & 99.85 &  99.96 & 99.93{\scriptsize$\pm$0.04} & 99.93{\scriptsize$\pm$0.04} & 99.93{\scriptsize$\pm$0.04} & 99.93{\scriptsize$\pm$0.04} \\
$P=2, Q=0$ (tanh)               & 99.81 & 100.00 & 99.95{\scriptsize$\pm$0.07} & 99.95{\scriptsize$\pm$0.07} & 99.95{\scriptsize$\pm$0.07} & 99.95{\scriptsize$\pm$0.07} \\
\textbf{\boldmath$P=2, Q=1$} (tanh) & 99.89 & 100.00 & 99.96{\scriptsize$\pm$0.04} & 99.96{\scriptsize$\pm$0.04} & 99.96{\scriptsize$\pm$0.04} & 99.96{\scriptsize$\pm$0.04} \\
$P=2, Q=1$ (LeakyReLU)          & 99.52 &  99.89 & 99.70{\scriptsize$\pm$0.14} & 99.70{\scriptsize$\pm$0.14} & 99.70{\scriptsize$\pm$0.14} & 99.70{\scriptsize$\pm$0.14} \\
$P=3, Q=0$ (tanh)               & 99.52 &  99.96 & 99.84{\scriptsize$\pm$0.17} & 99.85{\scriptsize$\pm$0.16} & 99.84{\scriptsize$\pm$0.17} & 99.84{\scriptsize$\pm$0.17} \\
\bottomrule
\end{tabularx}
\end{table*}

For the first accelerometer, located at the motor’s drive end, the 1D CNN $(P=1, Q=0)$ configuration with $\tanh$ activation function achieved an average test accuracy of 99.69\%~$\pm$~0.35\%. Replacing the activation function with $\mathrm{LeakyReLU}$ (with a negative slope of 0.01) across all convolutional layers led to a noticeable improvement, increasing the average accuracy to 99.74\%~$\pm$~0.22\%. Among the evaluated Self-ONN configurations, the model with $P=2$ and $Q=0$ attained the highest average test accuracy of 99.95\%~$\pm$~0.07\%, outperforming all CNN-based counterparts. However, the best overall fault diagnosis performance was achieved by the PadéNet model with $P=2$, $Q=1$ and $\tanh$ activation, which reached an average test accuracy of 99.96\%~$\pm$~0.04\% and an F1-score of 99.96\%~$\pm$~0.04\%. A detailed summary of the test metrics for all 1D CNN, Self-ONN, and PadéNet models evaluated using the first accelerometer data is provided in Table~\ref{tab_1staccel}.

Figure~\ref{fig5} presents the aggregated confusion matrices across 5 independent runs for the 1D CNN $(P=1, Q=0)$, Self-ONN $(P=2, Q=0)$, and PadéNet $(P=2, Q=1)$ models, using data from the first accelerometer as input. While all models achieve an average classification accuracy above 99\%, the 1D Self-ONN $(P=2, Q=0)$ and PadéNet $(P=2, Q=1)$ models demonstrate superior recall across all fault classes compared to 1D CNN $(P=1, Q=0)$. The Self-ONN $(P=2, Q=0)$ delivers similar performance to PadéNet $(P=2, Q=1)$ with fewer parameters, likely due to the placement of the first accelerometer at the motor’s drive end, which enables easier fault diagnosis through more effective feature extraction.

\begin{table*}[htbp]
\centering
\caption{Average test performance metrics (over 5 seeds) for different Padé models using the second accelerometer data as input.}
\label{tab_2ndaccel}
\renewcommand{\arraystretch}{1.1}
\setlength{\tabcolsep}{5pt}
\begin{tabularx}{\linewidth}{l *{6}{>{\centering\arraybackslash}X}}
\toprule
\textbf{Padé Model} & \textbf{Min. Acc. (\%)} & \textbf{Max. Acc. (\%)} & \textbf{Avg. Acc. (\%)} & \textbf{Avg. Prec. (\%)} & \textbf{Avg. Rec. (\%)} & \textbf{Avg. F1 (\%)} \\
\midrule
$P=1, Q=0$ (tanh)               & 92.75 & 94.57 & 93.94{\scriptsize$\pm$0.64} & 93.96{\scriptsize$\pm$0.63} & 93.94{\scriptsize$\pm$0.64} & 93.93{\scriptsize$\pm$0.64} \\
$P=1, Q=0$ (LeakyReLU)          & 96.06 & 97.62 & 96.81{\scriptsize$\pm$0.51} & 96.83{\scriptsize$\pm$0.49} & 96.81{\scriptsize$\pm$0.51} & 96.80{\scriptsize$\pm$0.50} \\
$P=1, Q=1$ (tanh)               & 93.79 & 95.35 & 94.79{\scriptsize$\pm$0.53} & 94.82{\scriptsize$\pm$0.54} & 94.79{\scriptsize$\pm$0.53} & 94.79{\scriptsize$\pm$0.53} \\
$P=1, Q=1$ (LeakyReLU)          & 96.61 & 97.88 & 97.25{\scriptsize$\pm$0.43} & 97.26{\scriptsize$\pm$0.43} & 97.25{\scriptsize$\pm$0.43} & 97.25{\scriptsize$\pm$0.43} \\
$P=1, Q=2$ (tanh)               & 95.05 & 96.28 & 95.47{\scriptsize$\pm$0.43} & 95.47{\scriptsize$\pm$0.43} & 95.47{\scriptsize$\pm$0.43} & 95.46{\scriptsize$\pm$0.43} \\
$P=1, Q=2$ (LeakyReLU)          & 96.95 & 97.81 & 97.54{\scriptsize$\pm$0.32} & 97.55{\scriptsize$\pm$0.33} & 97.54{\scriptsize$\pm$0.32} & 97.54{\scriptsize$\pm$0.32} \\
$P=2, Q=0$ (tanh)               & 95.31 & 96.24 & 95.96{\scriptsize$\pm$0.33} & 95.97{\scriptsize$\pm$0.32} & 95.96{\scriptsize$\pm$0.33} & 95.96{\scriptsize$\pm$0.33} \\
$P=2, Q=1$ (tanh)               & 96.35 & 97.25 & 96.85{\scriptsize$\pm$0.29} & 96.86{\scriptsize$\pm$0.28} & 96.85{\scriptsize$\pm$0.29} & 96.84{\scriptsize$\pm$0.29} \\
\textbf{\boldmath$P=2, Q=1$} (LeakyReLU) & 97.88 & 98.59 & 98.26{\scriptsize$\pm$0.26} & 98.27{\scriptsize$\pm$0.26} & 98.26{\scriptsize$\pm$0.26} & 98.26{\scriptsize$\pm$0.26} \\
$P=3, Q=0$ (tanh)               & 96.13 & 97.21 & 96.67{\scriptsize$\pm$0.34} & 96.68{\scriptsize$\pm$0.35} & 96.67{\scriptsize$\pm$0.34} & 96.67{\scriptsize$\pm$0.35} \\
\bottomrule
\end{tabularx}
\end{table*}

Located on the bearing housing near the drive end, the second accelerometer presented a complex diagnostic environment due to potential signal disturbances from bearing dynamics. When the second accelerometer data was used as input, the 1D CNN model ($P=1$, $Q=0$) with $\tanh$ activation reached an average test accuracy of 93.94\% $\pm$ 0.64\% as given in Table~\ref{tab_2ndaccel}. Switching to $\mathrm{LeakyReLU}$ activation across its convolutional layers significantly enhanced performance, boosting the average accuracy to 96.81\% $\pm$ 0.51\%, with reduced variability indicating improved consistency across each run. Among Self-ONN models tested, the configuration ($P=3$, $Q=0$) delivered the highest mean accuracy of 96.67\% $\pm$ 0.34\%. Although the average classification accuracy of the Self-ONN model ($P=3$, $Q=0$) surpassed that of the 1D CNN with $\tanh$ activation, switching to $\mathrm{LeakyReLU}$ enabled the 1D CNN to achieve higher fault diagnosis accuracy. Thus, it can be concluded that Self-ONNs generally outperform CNNs when $\tanh$ is employed as the activation function in both models. However, when $\mathrm{LeakyReLU}$ is used in all convolutional layers of a 1D CNN, it can achieve higher fault diagnosis accuracy compared to Self-ONNs utilizing the $\tanh$ activation function, potentially due to the vanishing gradient issue commonly associated with $\tanh$. In contrast to other 1D CNN and Self-ONN models, the most effective fault diagnosis for the second accelerometer was achieved by the 1D PadéNet model with $P=2, Q=1$ and $\mathrm{LeakyReLU}$ activation, having a mean test accuracy of 98.26\%~$\pm$~0.26\%, along with an F1-score of 98.26\%~$\pm$~0.26\%. A complete summary of performance metrics for all tested 1D CNN, Self-ONN, and PadéNet models using the second accelerometer data is available in Table~\ref{tab_2ndaccel}.

Figure~\ref{fig6} shows the aggregated confusion matrices across 5 independent runs for the 1D CNN $(P=1, Q=0)$, Self-ONN $(P=3, Q=0)$, and PadéNet $(P=2, Q=1)$ models, using data from the second accelerometer as input. For these confusion matrices, the 1D Self-ONN model uses the $\tanh$ activation function, while the 1D CNN and PadéNet utilize $\mathrm{LeakyReLU}$ with negative slope of 0.01. The 1D CNN $(P=1, Q=0)$ and Self-ONN $(P=3, Q=0)$ models achieve classification accuracies of 94.82\% and 93.75\%, respectively, in the faulty bearings (FB) class. In comparison, the 1D PadéNet $(P=2, Q=1)$ model delivers a significantly higher accuracy of 96.78\% in the same class. Similarly, for the rotor misalignment (RM) class, the 1D CNN and Self-ONN models yield accuracies of 93.63\% and 94.70\%, respectively, whereas the 1D PadéNet $(P=2, Q=1)$ model once again outperforms them with a superior accuracy of 96.67\%. Faulty bearings and rotor misalignment are two of the most critical fault types in electric motors, as they can severely compromise mechanical integrity and lead to costly operational downtimes if not detected early. Accurate classification of these faults is therefore essential for timely maintenance and fault prevention. The enhanced fault diagnosis performance of the 1D PadéNet model is clearly evident across all fault classes in these confusion matrices and reflects its superior capability to distinguish even the most critical fault types.

\begin{table*}[htbp]
\centering
\caption{Average test performance metrics (over 5 seeds) for different Padé models using the third accelerometer data as input.}
\label{tab_3rdaccel}
\renewcommand{\arraystretch}{1.1}
\setlength{\tabcolsep}{5pt}
\begin{tabularx}{\linewidth}{l *{6}{>{\centering\arraybackslash}X}}
\toprule
\textbf{Padé Model} & \textbf{Min. Acc. (\%)} & \textbf{Max. Acc. (\%)} & \textbf{Avg. Acc. (\%)} & \textbf{Avg. Prec. (\%)} & \textbf{Avg. Rec. (\%)} & \textbf{Avg. F1 (\%)} \\
\midrule
$P=1, Q=0$ (tanh)               & 93.12 & 94.38 & 93.74{\scriptsize$\pm$0.45} & 93.76{\scriptsize$\pm$0.44} & 93.74{\scriptsize$\pm$0.45} & 93.73{\scriptsize$\pm$0.44} \\
$P=1, Q=0$ (LeakyReLU)          & 96.35 & 97.40 & 96.73{\scriptsize$\pm$0.40} & 96.74{\scriptsize$\pm$0.40} & 96.73{\scriptsize$\pm$0.40} & 96.72{\scriptsize$\pm$0.40} \\
$P=1, Q=1$ (tanh)               & 93.34 & 94.61 & 94.11{\scriptsize$\pm$0.50} & 94.12{\scriptsize$\pm$0.50} & 94.11{\scriptsize$\pm$0.50} & 94.10{\scriptsize$\pm$0.50} \\
$P=1, Q=1$ (LeakyReLU)          & 96.99 & 97.62 & 97.30{\scriptsize$\pm$0.28} & 97.32{\scriptsize$\pm$0.28} & 97.30{\scriptsize$\pm$0.28} & 97.30{\scriptsize$\pm$0.28} \\
$P=1, Q=2$ (tanh)               & 93.68 & 95.76 & 94.81{\scriptsize$\pm$0.73} & 94.83{\scriptsize$\pm$0.73} & 94.81{\scriptsize$\pm$0.73} & 94.80{\scriptsize$\pm$0.74} \\
\textbf{\boldmath$P=1, Q=2$} (LeakyReLU) & 97.10 & 97.88 & 97.61{\scriptsize$\pm$0.27} & 97.62{\scriptsize$\pm$0.27} & 97.61{\scriptsize$\pm$0.27} & 97.61{\scriptsize$\pm$0.27} \\
$P=2, Q=0$ (tanh)               & 92.63 & 95.46 & 94.41{\scriptsize$\pm$1.01} & 94.43{\scriptsize$\pm$0.99} & 94.41{\scriptsize$\pm$1.01} & 94.41{\scriptsize$\pm$1.01} \\
$P=2, Q=1$ (tanh)               & 95.39 & 96.39 & 96.00{\scriptsize$\pm$0.36} & 96.03{\scriptsize$\pm$0.35} & 96.00{\scriptsize$\pm$0.36} & 96.00{\scriptsize$\pm$0.36} \\
$P=2, Q=1$ (LeakyReLU)          & 97.28 & 98.03 & 97.59{\scriptsize$\pm$0.27} & 97.61{\scriptsize$\pm$0.27} & 97.59{\scriptsize$\pm$0.27} & 97.59{\scriptsize$\pm$0.27} \\
$P=3, Q=0$ (tanh)               & 94.72 & 96.39 & 95.83{\scriptsize$\pm$0.59} & 95.87{\scriptsize$\pm$0.55} & 95.83{\scriptsize$\pm$0.59} & 95.83{\scriptsize$\pm$0.58} \\
\bottomrule
\end{tabularx}
\end{table*}

When data from the third accelerometer, positioned on the bearing housing farthest from the drive end, is used as input, the results in Table~\ref{tab_3rdaccel} show that the 1D PadéNet model, particularly with \( P=1 \), \( Q=2 \) and $\mathrm{LeakyReLU}$ activation, consistently outperforms both the 1D CNN and Self-ONN models across all evaluated metrics. In particular, the 1D PadéNet model achieves the highest average test accuracy of 97.61\% $\pm$ 0.27\%, surpassing the best-performing configurations of the 1D CNN and Self-ONN models. 

Figure~\ref{fig7} shows the aggregated confusion matrices for the 1D CNN $(P=1, Q=0)$, Self-ONN $(P=3, Q=0)$, and PadéNet $(P=1, Q=2)$ models, using data from the third accelerometer as input. The 1D PadéNet model once again achieves the highest per-class accuracies across all fault categories. Overall, although the classification accuracy of the 1D PadéNet model slightly decreases as the input accelerometer sensor is positioned farther from the drive-end, the 1D PadéNet consistently outperforms both the 1D CNN and Self-ONN models.

\begin{table*}[htbp]
\centering
\caption{Average test performance metrics (over 5 seeds) for different Padé models using acoustic data as input.}
\label{tab_audio}
\renewcommand{\arraystretch}{1.1}
\setlength{\tabcolsep}{5pt}
\begin{tabularx}{\linewidth}{l *{6}{>{\centering\arraybackslash}X}}
\toprule
\textbf{Padé Model} & \textbf{Min. Acc. (\%)} & \textbf{Max. Acc. (\%)} & \textbf{Avg. Acc. (\%)} & \textbf{Avg. Prec. (\%)} & \textbf{Avg. Rec. (\%)} & \textbf{Avg. F1 (\%)} \\
\midrule
$P=1, Q=0$ (tanh)               & 95.05 & 96.24 & 95.56{\scriptsize$\pm$0.45} & 95.58{\scriptsize$\pm$0.46} & 95.56{\scriptsize$\pm$0.45} & 95.55{\scriptsize$\pm$0.45} \\
$P=1, Q=0$ (LeakyReLU)          & 96.80 & 98.03 & 97.55{\scriptsize$\pm$0.46} & 97.57{\scriptsize$\pm$0.45} & 97.55{\scriptsize$\pm$0.46} & 97.55{\scriptsize$\pm$0.46} \\
$P=1, Q=1$ (tanh)               & 96.84 & 97.47 & 97.13{\scriptsize$\pm$0.24} & 97.14{\scriptsize$\pm$0.25} & 97.13{\scriptsize$\pm$0.24} & 97.13{\scriptsize$\pm$0.24} \\
$P=1, Q=1$ (LeakyReLU)          & 97.58 & 98.36 & 97.89{\scriptsize$\pm$0.30} & 97.92{\scriptsize$\pm$0.29} & 97.89{\scriptsize$\pm$0.30} & 97.89{\scriptsize$\pm$0.31} \\
$P=1, Q=2$ (tanh)               & 97.47 & 97.66 & 97.58{\scriptsize$\pm$0.07} & 97.59{\scriptsize$\pm$0.07} & 97.58{\scriptsize$\pm$0.07} & 97.58{\scriptsize$\pm$0.07} \\
\textbf{\boldmath$P=1, Q=2$} (LeakyReLU) & 97.81 & 98.92 & 98.33{\scriptsize$\pm$0.44} & 98.34{\scriptsize$\pm$0.43} & 98.33{\scriptsize$\pm$0.44} & 98.33{\scriptsize$\pm$0.44} \\
$P=2, Q=0$ (tanh)               & 97.62 & 98.10 & 97.89{\scriptsize$\pm$0.16} & 97.89{\scriptsize$\pm$0.16} & 97.89{\scriptsize$\pm$0.16} & 97.89{\scriptsize$\pm$0.16} \\
$P=2, Q=1$ (tanh)               & 97.02 & 98.62 & 98.07{\scriptsize$\pm$0.55} & 98.08{\scriptsize$\pm$0.54} & 98.07{\scriptsize$\pm$0.55} & 98.07{\scriptsize$\pm$0.55} \\
$P=2, Q=1$ (LeakyReLU)          & 97.92 & 98.70 & 98.30{\scriptsize$\pm$0.32} & 98.31{\scriptsize$\pm$0.31} & 98.30{\scriptsize$\pm$0.32} & 98.30{\scriptsize$\pm$0.32} \\
$P=3, Q=0$ (tanh)               & 97.51 & 98.40 & 97.93{\scriptsize$\pm$0.32} & 97.94{\scriptsize$\pm$0.32} & 97.93{\scriptsize$\pm$0.32} & 97.93{\scriptsize$\pm$0.32} \\
\bottomrule
\end{tabularx}
\end{table*}

When acoustic signals captured by the microphone are used as input, 1D PadéNet models continue to demonstrate superior diagnostic capabilities, as shown in Table~\ref{tab_audio}. The best performance was achieved by the configuration with \( P=1, Q=2 \) and $\mathrm{LeakyReLU}$ activation, which yielded the highest average accuracy of 98.33\%~$\pm$~0.44\%. This result is superior to those obtained using data from the second and third accelerometers, yet slightly underperforms compared to the first accelerometer, which is mounted directly above the drive-end.

Figure~\ref{fig8} shows the aggregated confusion matrices across 5 independent runs for the 1D CNN $(P=1, Q=0)$, Self-ONN $(P=3, Q=0)$, and PadéNet $(P=1, Q=2)$ models, using acoustic data as input. The PadéNet model with $P=1$ and $Q=2$ achieves the highest per-class classification accuracies in the BR, FB, RM, and RU fault categories, while the Self-ONN model performs best in the H, KA, and VU classes. The 1D CNN model achieves the highest accuracy only in the SW category. As a result, acoustic data can serve as a highly informative and reliable modality for fault diagnosis when processed with the 1D PadéNet architecture, and we can obtain a diagnostic performance comparable to accelerometer-based inputs.

\begin{table*}[htbp] 
\caption{Comparison of DL-based methods for fault diagnosis using the University of Ottawa's constant speed vibration and acoustic datasets under unloaded and loaded conditions. \label{tab_literature}}
\begin{tabularx}{\linewidth}{cccc}
\toprule
\textbf{Model}	& \textbf{Sensor}	& \textbf{Average Test Accuracy (\%)}	& \textbf{No. of Params.} \\
\midrule
13-Layers CNN \citep{ertargin2024mechanical} & Accelerometer-1 & 97.73 & - \\
15-Layers CNN \citep{ertargin2024mechanical} & Accelerometer-1 & 99.44 & - \\
CNN-LSTM \citep{ertargin2024mechanical} & Accelerometer-1 & 99.96 & 246,568 \\
$P=1, Q=0$ (CNN) \textsuperscript{1} & Accelerometer-1 & 99.74 & 58,376 \\
$P=2, Q=0$ (Self-ONN) \textsuperscript{2} & Accelerometer-1 & 99.95 & 101,832 \\
$P=2, Q=1$ (PadéNet) \textsuperscript{3} & Accelerometer-1 & 99.96 & 145,064 \\
\midrule
13-Layers CNN \citep{ertargin2024mechanical} & Accelerometer-2 & 80.58 & - \\
15-Layers CNN \citep{ertargin2024mechanical} & Accelerometer-2 & 83.44 & - \\
CNN-LSTM \citep{ertargin2024mechanical} & Accelerometer-2 & 98.88 & 246,568 \\
$P=1, Q=0$ (CNN) \textsuperscript{1} & Accelerometer-2 & 96.81 & 58,376 \\
$P=3, Q=0$ (Self-ONN) \textsuperscript{2} & Accelerometer-2 & 96.67 & 145,288 \\
$P=2, Q=1$ (PadéNet) \textsuperscript{3} & Accelerometer-2 & 98.26 & 145,064 \\
\midrule
13-Layers CNN \citep{ertargin2024mechanical} & Accelerometer-3 & 82.89 & - \\
15-Layers CNN \citep{ertargin2024mechanical} & Accelerometer-3 & 87.43 & - \\
CNN-LSTM \citep{ertargin2024mechanical} & Accelerometer-3 & 99.37 & 246,568 \\
$P=1, Q=0$ (CNN) \textsuperscript{1} & Accelerometer-3 & 96.73 & 58,376 \\
$P=3, Q=0$ (Self-ONN) \textsuperscript{2} & Accelerometer-3 & 95.83 & 145,288 \\
$P=1, Q=2$ (PadéNet) \textsuperscript{3} & Accelerometer-3 & 97.61 & 144,840 \\
\midrule
Custom 2D CNN \citep{ertarginclassifying} & Microphone & 83.36 & - \\
VGG16 \citep{ertarginclassifying} & Microphone & 91.52 & 134,293,320 \\
VGG19 \citep{ertarginclassifying} & Microphone & 92.11 & 139,602,016 \\
CNN-LSTM \citep{ertargin2024mechanical} & Microphone & 96.55 & 246,568 \\
$P=1, Q=0$ (CNN) \textsuperscript{1} & Microphone & 97.55 & 58,376 \\
$P=3, Q=0$ (Self-ONN) \textsuperscript{2} & Microphone & 97.93 & 145,288 \\
$P=1, Q=2$ (PadéNet) \textsuperscript{3} & Microphone & 98.33 & 144,840 \\
\bottomrule
\end{tabularx}
\noindent{\footnotesize
\textsuperscript{1} The best-performing 1D CNN models based on average classification accuracy. \\
\textsuperscript{2} The best-performing 1D Self-ONN models based on average classification accuracy. \\
\textsuperscript{3} The best-performing 1D PadéNet models based on average classification accuracy.}
\end{table*}

The fault diagnosis performance of 1D PadéNets is also compared with other DL-based methods trained on the University of Ottawa’s constant-speed vibration and audio datasets. In \citep{ertargin2024mechanical}, a 15-layer 1D CNN-LSTM model (comprising 6 Conv1D and 2 LSTM layers) was proposed for classifying electrical and mechanical faults. The model takes an input window size of 1000 samples, as used in this study, with the same train-validation-test split ratios and all accelerometer inputs. To evaluate the impact of the LSTM layers, two other variations of the model were tested. First, the two LSTM layers were removed, resulting in a 13-layer CNN, referred to as “13-Layers CNN” in Table~\ref{tab_literature}. In the second experiment, the LSTM layers in the 11th and 12th layers were replaced with Conv1D layers, producing a 15-layer CNN. All results are presented in Table~\ref{tab_literature} for each accelerometer sensor. For Accelerometer-1, the 1D PadéNet ($P=2$, $Q=1$) achieves the same average test accuracy as the 1D CNN-LSTM model but with significantly fewer trainable parameters, and outperforms all other CNN models. Moreover, for accelerometers 2 and 3, 1D PadéNet delivers comparable classification accuracy to the CNN-LSTM model, while the 13-layer and 15-layer CNN models fail to match those accuracies. However, it is worth noting that, due to the sequential nature of LSTM layers, the CNN-LSTM model tends to be slower during training and inference, whereas the 1D PadéNet achieves similar performance with greater computational efficiency.

For the acoustic modality, we also compare classification performance against established baselines. In particular, we re-implemented the 1D CNN--LSTM from \citep{ertargin2024mechanical}, training it end-to-end using the same segmentation, normalization pipeline, and data splits as PadéNet to ensure a fair comparison. 
As shown in Table~\ref{tab_literature}, PadéNet with ($P=1$, $Q=2$) achieves an average test accuracy of $98.33\%$ with $144,840$ parameters, outperforming both the 1D CNN ($97.55\%$, $58,376$ parameters) and the CNN-LSTM ($96.55\%$, $246,568$ parameters). Moreover, in \citep{ertarginclassifying}, acoustic signals from the Ottawa University constant-speed audio dataset were converted into spectrograms, and faults in induction motors were detected using a transfer learning approach with pre-trained models. An 8-class fault detection task was performed, achieving an accuracy rate of 91.52\% using the VGG16 model and 92.11\% using the VGG19 model. In the same study, a custom 2D CNN consisting of 4 convolutional, 4 max-pooling, and 3 dense layers was also evaluated for comparison. The 1D PadéNet ($P=1$, $Q=2$) outperforms these models, achieving a superior fault diagnosis performance with approximately a 6\% increase in accuracy, while requiring far fewer parameters than the pretrained models. Consequently, 1D PadéNets are capable of achieving high classification accuracies using raw audio data as input, while requiring significantly fewer parameters.

\subsection{Sensitivity Analysis} \label{sec4_2}
\begin{figure*}[htbp]
\includegraphics[width=\textwidth]{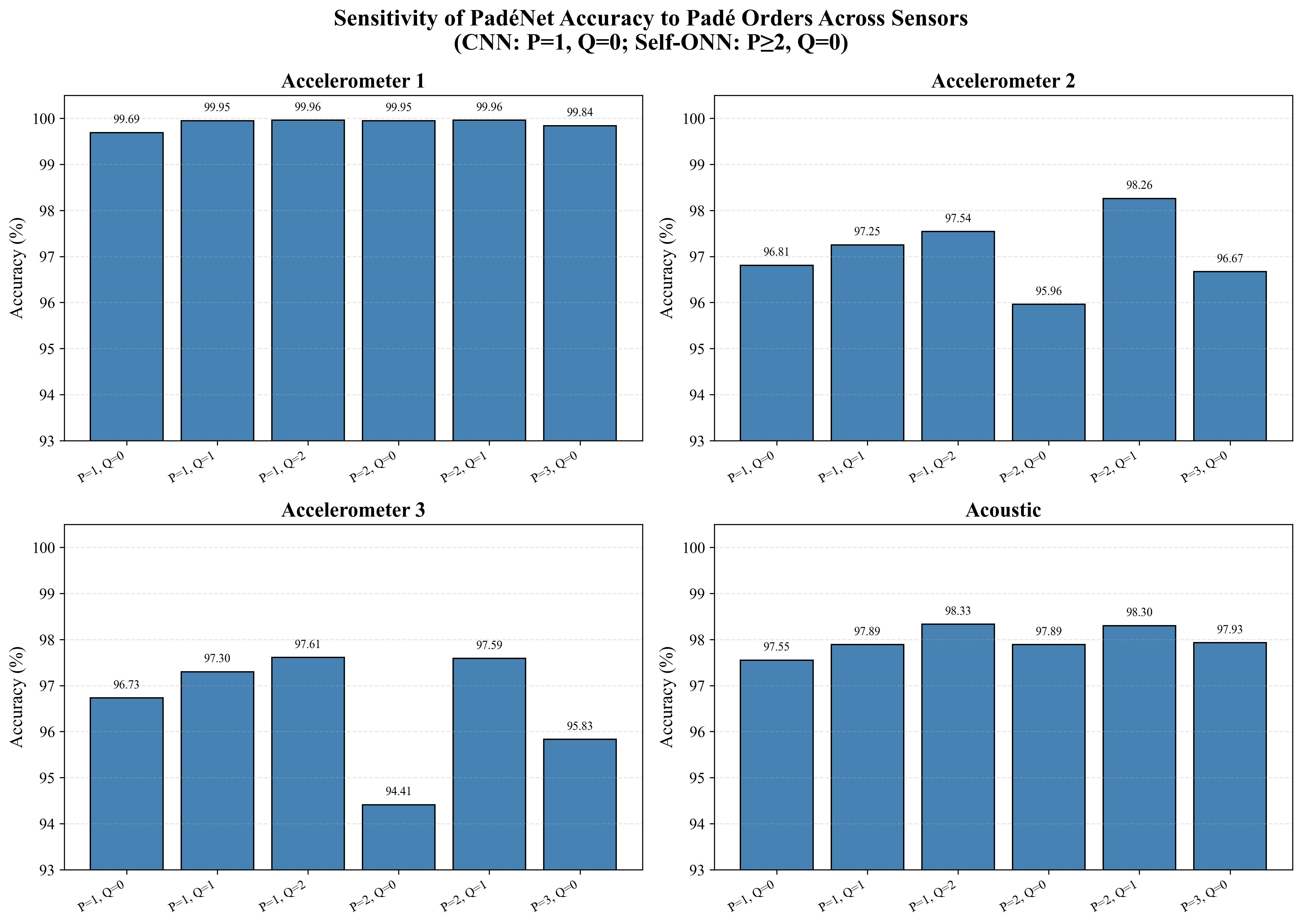}
\caption{Sensitivity of PadéNet accuracy to Padé orders across sensors.
\label{fig9}}
\end{figure*}

\begin{figure*}[htbp]
\includegraphics[width=\textwidth]{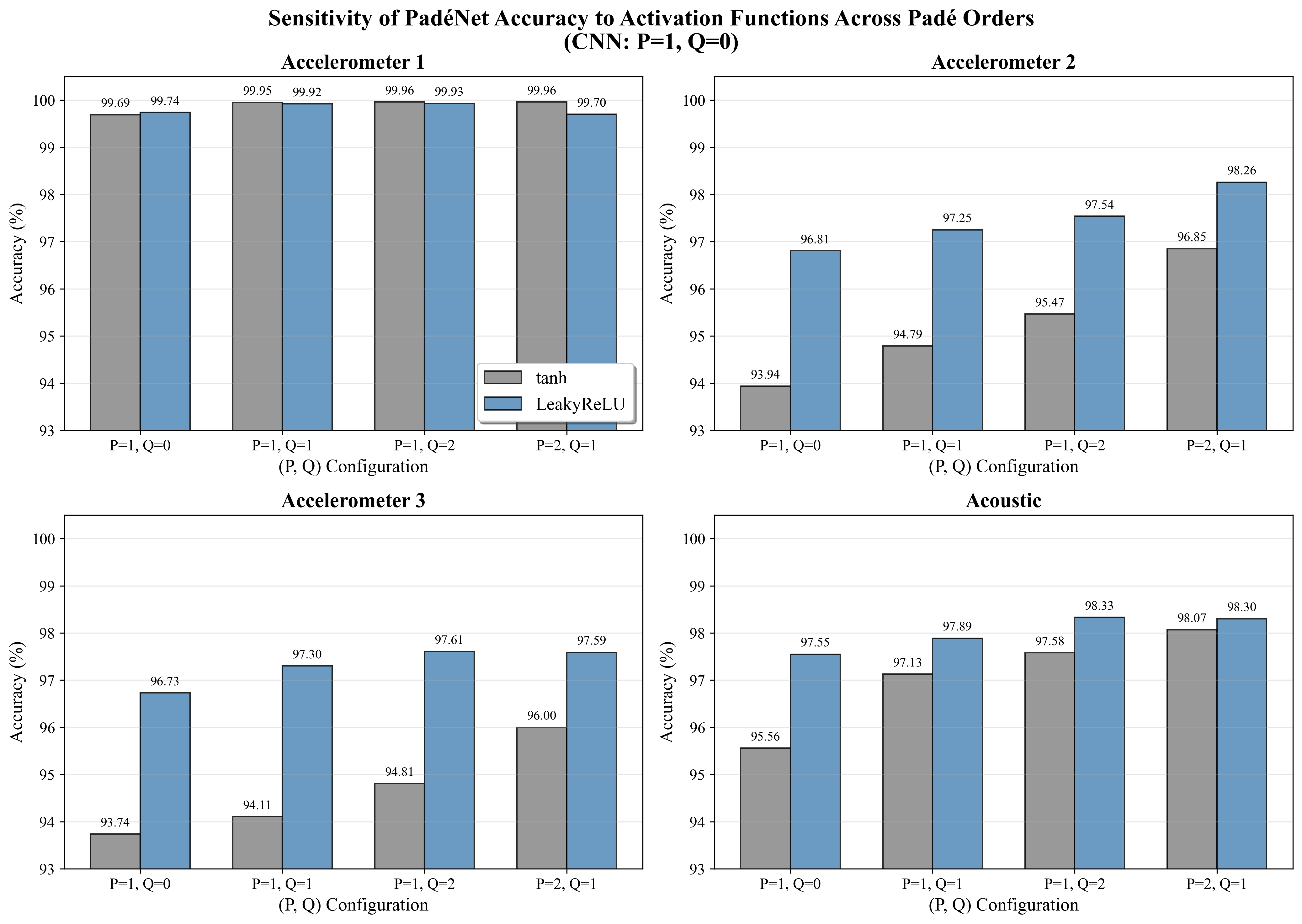}
\caption{Sensitivity of PadéNet accuracy to activation functions across Padé orders.
\label{fig10}}
\end{figure*}

We also provide an explicit sensitivity analysis across Padé orders and activation functions on identical train/validation/test splits for all sensors. Figure~\ref{fig9} summarizes average test accuracies for $P\in\{1,2,3\}$ and $Q\in\{0,1,2\}$ (per sensor). Figure~\ref{fig10} examines the interaction with the activation function ($\tanh$ versus $\mathrm{LeakyReLU}$).

The dominant factor is the inclusion of a denominator branch ($Q>0$). Adding a modest convolutional denominator in the Padé formulation ($Q=1$ or $2$) consistently improves accuracy over both the CNN baseline and the Self-ONN family, as it stabilizes higher-order terms while preserving expressivity. For example, accuracy on Accelerometer-2 increases from 96.81\% with CNN $(P{=}1,\,Q{=}0)$ to 98.26\% with PadéNet $(P{=}2,\,Q{=}1)$; on Accelerometer-3 from 96.73\% with CNN $(P{=}1,\,Q{=}0)$ to 97.61\% with PadéNet $(P{=}1,\,Q{=}2)$; and on the acoustic sensor from 97.55\% with CNN $(P{=}1,\,Q{=}0)$ to 98.33\% with PadéNet $(P{=}1,\,Q{=}2)$ (Figure~\ref{fig9}). Accelerometer-1 is already at a performance plateau but still rises from 99.69\% with CNN $(P{=}1,\,Q{=}0)$ to approximately 99.96\% with any PadéNet having $Q>0$. Increasing the numerator order without a denominator (i.e., Self-ONN with $Q{=}0$ and $P{\ge}2$) does not reliably help and can sometimes reduce accuracy; on Accelerometer-3, Self-ONN $(P{=}2,\,Q{=}0)$ drops to 94.41\% (Figure~\ref{fig9}). Overall, balanced low-order configurations such as PadéNet $(P{=}2,\,Q{=}1)$ or $(P{=}1,\,Q{=}2)$ consistently provide the best trade-off between accuracy, training stability, and computational complexity.

Across the configurations we tested, $\mathrm{LeakyReLU}$ generally improves accuracy over $\tanh$ for nearly all $(P,Q)$ pairs where training is stable. For the CNN baseline $(P{=}1,Q{=}0)$, $\mathrm{LeakyReLU}$ raises accuracy from 93.94\% to 96.81\% on Accelerometer-2, from 93.74\% to 96.73\% on Accelerometer-3, and from 95.56\% to 97.55\% on the acoustic sensor (Figure~\ref{fig10}). Accelerometer-1 is already at a performance plateau, remaining around 99.7--99.8\% with either activation. For Self-ONN $(Q{=}0, P{\ge}2)$, unbounded activations cannot be used reliably, so comparisons are limited to $\tanh$; moreover, increasing the polynomial order may not improve accuracy, as noted above. For PadéNet $(P{>}0, Q{>}0)$, the learned denominator stabilizes higher-order terms and enables the safe use of unbounded activations; with $\mathrm{LeakyReLU}$, PadéNet $(P{=}2,Q{=}1)$ on Accelerometer-2 and PadéNet $(P{=}1,Q{=}2)$ on Accelerometer-3 and on the acoustic sensor achieve the strongest results, while on Accelerometer-1 both activations are effectively tied near 99.96\%.

\begin{table*}[htbp]
\centering
\scriptsize
\setlength{\tabcolsep}{3pt} 
\caption{Average classification accuracies (mean $\pm$ s.d.) over 5 runs across SNR levels using Accelerometer-1 input.}
\label{tab:acc1_snr}
\resizebox{\linewidth}{!}{%
\begin{tabular}{llcccccccc}
\toprule
\multicolumn{2}{c}{} & \multicolumn{8}{c}{\textbf{SNR (dB)}} \\
\cmidrule(lr){3-10}
\textbf{Model} & \textbf{Activation} & \textbf{-4} & \textbf{-2} & \textbf{0} & \textbf{2} & \textbf{4} & \textbf{6} & \textbf{8} & \textbf{10} \\
\midrule
P=1, Q=0 & tanh & 52.86 $\pm$ 1.23 & 65.19 $\pm$ 0.82 & 75.47 $\pm$ 0.75 & 83.50 $\pm$ 0.65 & 88.59 $\pm$ 0.66 & 91.89 $\pm$ 0.41 & 94.07 $\pm$ 0.62 & 95.30 $\pm$ 0.50 \\
P=1, Q=0 & LeakyReLU & 63.44 $\pm$ 1.48 & 77.43 $\pm$ 1.16 & 86.85 $\pm$ 0.68 & 92.70 $\pm$ 0.58 & 95.65 $\pm$ 0.52 & 97.24 $\pm$ 0.53 & 98.15 $\pm$ 0.31 & 98.68 $\pm$ 0.13 \\
P=1, Q=1 & tanh & 56.70 $\pm$ 0.48 & 69.42 $\pm$ 0.81 & 79.90 $\pm$ 0.69 & 87.66 $\pm$ 0.50 & 92.06 $\pm$ 0.50 & 94.64 $\pm$ 0.49 & 96.43 $\pm$ 0.74 & 97.54 $\pm$ 0.61 \\
\textbf{P=1, Q=1} & \textbf{LeakyReLU} & \textbf{64.93 $\pm$ 0.99} & \textbf{78.80 $\pm$ 0.61} & \textbf{88.33 $\pm$ 0.66} & \textbf{93.47 $\pm$ 0.37} & \textbf{96.19 $\pm$ 0.36} & \textbf{97.67 $\pm$ 0.32} & \textbf{98.49 $\pm$ 0.18} & \textbf{98.91 $\pm$ 0.16} \\
P=1, Q=2 & tanh & 58.52 $\pm$ 0.54 & 71.72 $\pm$ 0.44 & 82.71 $\pm$ 0.54 & 89.67 $\pm$ 0.29 & 93.47 $\pm$ 0.26 & 95.65 $\pm$ 0.29 & 96.93 $\pm$ 0.15 & 97.83 $\pm$ 0.14 \\
P=1, Q=2 & LeakyReLU & 63.71 $\pm$ 1.76 & 77.54 $\pm$ 0.77 & 87.37 $\pm$ 0.63 & 93.12 $\pm$ 0.41 & 95.82 $\pm$ 0.24 & 97.31 $\pm$ 0.38 & 98.25 $\pm$ 0.26 & 98.76 $\pm$ 0.20 \\
P=2, Q=0 & tanh & 54.55 $\pm$ 2.65 & 67.43 $\pm$ 2.43 & 78.39 $\pm$ 2.03 & 86.39 $\pm$ 1.50 & 90.99 $\pm$ 0.75 & 93.55 $\pm$ 0.83 & 95.34 $\pm$ 0.76 & 96.40 $\pm$ 0.93 \\
P=2, Q=1 & tanh & 59.05 $\pm$ 1.63 & 72.02 $\pm$ 0.99 & 82.99 $\pm$ 0.55 & 89.61 $\pm$ 0.32 & 93.74 $\pm$ 0.31 & 95.86 $\pm$ 0.35 & 97.34 $\pm$ 0.09 & 98.07 $\pm$ 0.14 \\
P=2, Q=1 & LeakyReLU & 64.38 $\pm$ 0.87 & 77.63 $\pm$ 1.57 & 86.85 $\pm$ 1.21 & 92.70 $\pm$ 0.67 & 95.77 $\pm$ 0.40 & 97.28 $\pm$ 0.28 & 98.32 $\pm$ 0.41 & 98.85 $\pm$ 0.31 \\
P=3, Q=0 & tanh & 58.18 $\pm$ 1.07 & 71.02 $\pm$ 0.60 & 81.54 $\pm$ 0.69 & 88.54 $\pm$ 0.77 & 92.86 $\pm$ 0.67 & 95.00 $\pm$ 0.71 & 96.55 $\pm$ 0.37 & 97.38 $\pm$ 0.47 \\
\bottomrule
\end{tabular}}
\end{table*}

\begin{table*}[htbp]
\centering
\scriptsize
\setlength{\tabcolsep}{3pt} 
\caption{Average classification accuracies (mean $\pm$ s.d.) over 5 runs across SNR levels using Accelerometer-2 input.}
\label{tab:acc2_snr}
\resizebox{\linewidth}{!}{%
\begin{tabular}{llcccccccc}
\toprule
\multicolumn{2}{c}{} & \multicolumn{8}{c}{\textbf{SNR (dB)}} \\
\cmidrule(lr){3-10}
\textbf{Model} & \textbf{Activation} & \textbf{-4} & \textbf{-2} & \textbf{0} & \textbf{2} & \textbf{4} & \textbf{6} & \textbf{8} & \textbf{10} \\
\midrule
P=1, Q=0 & tanh & 49.00 $\pm$ 1.18 & 63.75 $\pm$ 0.89 & 73.85 $\pm$ 1.05 & 80.32 $\pm$ 0.84 & 84.08 $\pm$ 0.75 & 86.42 $\pm$ 0.63 & 87.66 $\pm$ 0.71 & 88.50 $\pm$ 0.91 \\
P=1, Q=0 & LeakyReLU & 54.26 $\pm$ 4.80 & 77.13 $\pm$ 0.79 & 87.42 $\pm$ 0.75 & 91.89 $\pm$ 0.69 & 93.78 $\pm$ 0.70 & 94.64 $\pm$ 0.52 & 95.13 $\pm$ 0.64 & 95.43 $\pm$ 0.41 \\
P=1, Q=1 & tanh & 50.65 $\pm$ 0.95 & 66.69 $\pm$ 1.13 & 77.05 $\pm$ 1.32 & 82.83 $\pm$ 1.38 & 86.47 $\pm$ 1.43 & 88.53 $\pm$ 1.18 & 89.68 $\pm$ 1.31 & 90.49 $\pm$ 1.34 \\
P=1, Q=1 & LeakyReLU & 53.10 $\pm$ 2.90 & 76.62 $\pm$ 1.02 & 87.24 $\pm$ 0.47 & 92.03 $\pm$ 0.62 & 94.08 $\pm$ 0.58 & 95.13 $\pm$ 0.45 & 95.67 $\pm$ 0.40 & 95.82 $\pm$ 0.31 \\
P=1, Q=2 & tanh & 51.21 $\pm$ 2.59 & 68.42 $\pm$ 0.98 & 78.92 $\pm$ 0.76 & 85.01 $\pm$ 0.78 & 88.42 $\pm$ 0.63 & 90.09 $\pm$ 0.49 & 91.11 $\pm$ 0.49 & 91.84 $\pm$ 0.48 \\
P=1, Q=2 & LeakyReLU & 52.00 $\pm$ 4.67 & 76.12 $\pm$ 2.48 & 88.08 $\pm$ 0.97 & 92.84 $\pm$ 0.91 & 94.94 $\pm$ 0.66 & 95.69 $\pm$ 0.67 & 96.26 $\pm$ 0.61 & 96.44 $\pm$ 0.57 \\
P=2, Q=0 & tanh & 45.29 $\pm$ 2.35 & 64.55 $\pm$ 1.37 & 76.90 $\pm$ 0.34 & 83.59 $\pm$ 0.14 & 87.25 $\pm$ 0.35 & 89.35 $\pm$ 0.33 & 90.35 $\pm$ 0.15 & 91.15 $\pm$ 0.24 \\
P=2, Q=1 & tanh & 47.76 $\pm$ 1.25 & 68.17 $\pm$ 0.67 & 80.51 $\pm$ 0.82 & 86.96 $\pm$ 0.97 & 90.35 $\pm$ 0.60 & 92.19 $\pm$ 0.52 & 93.19 $\pm$ 0.33 & 93.82 $\pm$ 0.37 \\
\textbf{P=2, Q=1} & \textbf{LeakyReLU} & \textbf{55.77 $\pm$ 1.96} & \textbf{77.67 $\pm$ 2.16} & \textbf{88.92 $\pm$ 1.08} & \textbf{93.45 $\pm$ 0.92} & \textbf{95.20 $\pm$ 0.77} & \textbf{95.86 $\pm$ 0.85} & \textbf{96.29 $\pm$ 0.74} & \textbf{96.53 $\pm$ 0.61} \\
P=3, Q=0 & tanh & 44.49 $\pm$ 2.48 & 64.74 $\pm$ 1.32 & 77.82 $\pm$ 0.76 & 85.24 $\pm$ 0.66 & 89.26 $\pm$ 0.77 & 91.25 $\pm$ 0.86 & 92.29 $\pm$ 0.80 & 92.88 $\pm$ 0.72 \\\bottomrule
\end{tabular}}
\end{table*}

\begin{table*}[htbp]
\centering
\scriptsize
\setlength{\tabcolsep}{3pt} 
\caption{Average classification accuracies (mean $\pm$ s.d.) over 5 runs across SNR levels using Accelerometer-3 input.}
\label{tab:acc3_snr}
\resizebox{\linewidth}{!}{%
\begin{tabular}{llcccccccc}
\toprule
\multicolumn{2}{c}{} & \multicolumn{8}{c}{\textbf{SNR (dB)}} \\
\cmidrule(lr){3-10}
\textbf{Model} & \textbf{Activation} & \textbf{-4} & \textbf{-2} & \textbf{0} & \textbf{2} & \textbf{4} & \textbf{6} & \textbf{8} & \textbf{10} \\
\midrule
P=1, Q=0 & tanh & 58.01 $\pm$ 1.38 & 69.71 $\pm$ 1.02 & 78.30 $\pm$ 0.93 & 83.44 $\pm$ 1.13 & 86.10 $\pm$ 0.67 & 87.43 $\pm$ 0.56 & 88.17 $\pm$ 0.60 & 88.82 $\pm$ 0.52 \\
P=1, Q=0 & LeakyReLU & 64.50 $\pm$ 1.92 & 80.78 $\pm$ 0.91 & 88.56 $\pm$ 0.91 & 92.40 $\pm$ 0.58 & 93.87 $\pm$ 0.61 & 94.61 $\pm$ 0.53 & 94.81 $\pm$ 0.45 & 95.19 $\pm$ 0.34 \\
P=1, Q=1 & tanh & 59.43 $\pm$ 1.17 & 71.67 $\pm$ 0.84 & 80.23 $\pm$ 0.82 & 85.31 $\pm$ 1.08 & 87.60 $\pm$ 1.31 & 88.86 $\pm$ 1.48 & 89.75 $\pm$ 1.41 & 90.41 $\pm$ 1.33 \\
P=1, Q=1 & LeakyReLU & 65.98 $\pm$ 3.44 & 81.13 $\pm$ 2.38 & 89.41 $\pm$ 1.25 & 92.80 $\pm$ 0.77 & 94.55 $\pm$ 0.70 & 95.16 $\pm$ 0.72 & 95.68 $\pm$ 0.68 & 95.95 $\pm$ 0.72 \\
P=1, Q=2 & tanh & 60.15 $\pm$ 1.12 & 73.90 $\pm$ 1.23 & 83.72 $\pm$ 1.08 & 88.39 $\pm$ 0.80 & 90.89 $\pm$ 0.92 & 92.14 $\pm$ 0.95 & 92.93 $\pm$ 1.08 & 93.30 $\pm$ 1.04 \\
\textbf{P=1, Q=2} & \textbf{LeakyReLU} & \textbf{67.93 $\pm$ 1.93} & \textbf{82.69 $\pm$ 1.67} & \textbf{90.23 $\pm$ 1.25} & \textbf{93.57 $\pm$ 0.68} & \textbf{95.01 $\pm$ 0.38} & \textbf{95.73 $\pm$ 0.53} & \textbf{96.12 $\pm$ 0.24} & \textbf{96.29 $\pm$ 0.22} \\
P=2, Q=0 & tanh & 58.10 $\pm$ 2.84 & 70.68 $\pm$ 1.94 & 79.72 $\pm$ 1.01 & 84.84 $\pm$ 1.03 & 88.01 $\pm$ 0.81 & 89.58 $\pm$ 0.92 & 90.60 $\pm$ 0.89 & 90.97 $\pm$ 0.76 \\
P=2, Q=1 & tanh & 59.61 $\pm$ 1.44 & 72.94 $\pm$ 0.85 & 81.76 $\pm$ 0.59 & 86.56 $\pm$ 0.54 & 89.30 $\pm$ 0.68 & 90.58 $\pm$ 0.62 & 91.29 $\pm$ 0.64 & 91.49 $\pm$ 0.55 \\
P=2, Q=1 & LeakyReLU & 65.54 $\pm$ 1.20 & 81.27 $\pm$ 0.89 & 89.76 $\pm$ 0.51 & 93.50 $\pm$ 0.31 & 94.98 $\pm$ 0.75 & 95.62 $\pm$ 0.41 & 96.01 $\pm$ 0.53 & 96.21 $\pm$ 0.49 \\
P=3, Q=0 & tanh & 58.56 $\pm$ 1.07 & 73.12 $\pm$ 0.97 & 82.48 $\pm$ 0.96 & 88.07 $\pm$ 0.81 & 90.95 $\pm$ 0.68 & 92.46 $\pm$ 0.67 & 93.34 $\pm$ 0.66 & 93.84 $\pm$ 0.75 \\\bottomrule
\end{tabular}}
\end{table*}

\begin{table*}[htbp]
\centering
\scriptsize
\setlength{\tabcolsep}{3pt} 
\caption{Average classification accuracies (mean $\pm$ s.d.) over 5 runs across SNR levels using Microphone input.}
\label{tab:mic_snr}
\resizebox{\linewidth}{!}{%
\begin{tabular}{llcccccccc}
\toprule
\multicolumn{2}{c}{} & \multicolumn{8}{c}{\textbf{SNR (dB)}} \\
\cmidrule(lr){3-10}
\textbf{Model} & \textbf{Activation} & \textbf{-4} & \textbf{-2} & \textbf{0} & \textbf{2} & \textbf{4} & \textbf{6} & \textbf{8} & \textbf{10} \\
\midrule
P=1, Q=0 & tanh & 37.95 $\pm$ 0.42 & 46.27 $\pm$ 0.63 & 54.66 $\pm$ 0.61 & 61.55 $\pm$ 1.07 & 67.09 $\pm$ 0.98 & 71.45 $\pm$ 1.09 & 74.39 $\pm$ 1.21 & 76.91 $\pm$ 1.21 \\
P=1, Q=0 & LeakyReLU & 45.95 $\pm$ 1.42 & 61.18 $\pm$ 1.91 & 72.42 $\pm$ 1.77 & 78.97 $\pm$ 1.80 & 83.59 $\pm$ 1.84 & 86.96 $\pm$ 1.52 & 89.11 $\pm$ 1.07 & 90.18 $\pm$ 1.33 \\
P=1, Q=1 & tanh & 42.17 $\pm$ 1.85 & 52.22 $\pm$ 1.45 & 60.45 $\pm$ 1.09 & 67.45 $\pm$ 0.95 & 72.66 $\pm$ 0.97 & 76.79 $\pm$ 0.98 & 80.26 $\pm$ 1.01 & 82.43 $\pm$ 1.11 \\
P=1, Q=1 & LeakyReLU & 50.82 $\pm$ 2.26 & 62.47 $\pm$ 1.11 & 73.05 $\pm$ 1.25 & 79.67 $\pm$ 0.80 & 84.30 $\pm$ 0.94 & 87.59 $\pm$ 0.83 & 89.76 $\pm$ 0.60 & 91.24 $\pm$ 0.77 \\
P=1, Q=2 & tanh & 42.17 $\pm$ 1.99 & 52.64 $\pm$ 0.80 & 61.83 $\pm$ 0.76 & 68.52 $\pm$ 1.46 & 74.00 $\pm$ 1.25 & 78.24 $\pm$ 1.04 & 81.45 $\pm$ 1.26 & 83.90 $\pm$ 1.29 \\
\textbf{P=1, Q=2} & \textbf{LeakyReLU} & \textbf{52.07 $\pm$ 2.21} & \textbf{63.96 $\pm$ 0.51} & \textbf{73.42 $\pm$ 0.44} & \textbf{79.75 $\pm$ 1.66} & \textbf{84.38 $\pm$ 0.92} & \textbf{87.80 $\pm$ 0.69} & \textbf{90.10 $\pm$ 0.64} & \textbf{91.66 $\pm$ 0.53} \\
P=2, Q=0 & tanh & 36.86 $\pm$ 1.69 & 47.10 $\pm$ 1.04 & 56.78 $\pm$ 0.71 & 64.12 $\pm$ 0.66 & 70.53 $\pm$ 0.63 & 75.21 $\pm$ 0.48 & 78.80 $\pm$ 0.45 & 81.82 $\pm$ 0.70 \\
P=2, Q=1 & tanh & 38.72 $\pm$ 2.00 & 51.69 $\pm$ 1.88 & 62.98 $\pm$ 1.74 & 70.99 $\pm$ 1.39 & 76.27 $\pm$ 1.09 & 80.02 $\pm$ 1.25 & 83.31 $\pm$ 1.19 & 86.02 $\pm$ 1.38 \\
P=2, Q=1 & LeakyReLU & 47.93 $\pm$ 1.39 & 63.44 $\pm$ 1.51 & 72.33 $\pm$ 0.73 & 79.26 $\pm$ 0.82 & 84.38 $\pm$ 1.51 & 87.33 $\pm$ 1.31 & 89.13 $\pm$ 1.49 & 90.69 $\pm$ 1.51 \\
P=3, Q=0 & tanh & 35.42 $\pm$ 1.85 & 46.82 $\pm$ 1.62 & 57.81 $\pm$ 0.95 & 66.05 $\pm$ 0.71 & 71.93 $\pm$ 0.77 & 76.74 $\pm$ 0.70 & 80.26 $\pm$ 0.59 & 82.83 $\pm$ 0.57 \\\bottomrule
\end{tabular}}
\end{table*}

\subsection{Performance under Additive Gaussian Noise} \label{sec4_3}
We further extend our evaluation to include 1D CNNs, 1D Self-ONNs, and 1D PadéNets under varying noise conditions. The same data partitioning protocol described in Section~\ref{sec3_2} is adopted. For the training and validation sets, each clean segment is subjected to stochastic noise injection with a probability of \(p = 0.5\). When noise is injected, zero–mean Gaussian noise is added at a target signal–to–noise ratio (SNR) drawn uniformly at random from a predefined range of \(0\)–\(6\)~dB.  

For evaluation, noisy test sets are generated at fixed SNR levels by creating separate test copies for each target value in \(\{-4,\,-2,\,0,\,2,\,4,\,6,\,8,\,10\}\)~dB. In each case, noise is added on a per–segment basis according to:  

\begin{equation}
\begin{aligned}
\mathbf{x}_{\mathrm{noisy}} &= \mathbf{x}_{\mathrm{clean}} + \mathbf{n}, \\
\mathbf{n} &\sim \mathcal{N}\!\left(0,\sigma^2\right),\quad
\sigma = \sqrt{\frac{P_{\mathrm{signal}}}{10^{\mathrm{SNR}/10}}}
\end{aligned}
\label{eq:noise_addition}
\end{equation}

where \(P_{\mathrm{signal}} = \frac{1}{L} \sum_{i=1}^L x_{\mathrm{clean},i}^2\) denotes the average power of the clean segment of length \(L\), and \(\sigma^2\) is the noise variance required to achieve the desired SNR in decibels. By using distinct fixed SNR levels during testing that are not necessarily seen during training, we can also quantify each model’s ability to generalize to unseen noise conditions.

The comparative analysis in Tables~\ref{tab:acc1_snr}–\ref{tab:mic_snr} demonstrates that the proposed 1D PadéNet configurations consistently deliver superior classification accuracy relative to both the baseline 1D CNN and 1D Self-ONN models under a wide range of SNR conditions. In particular, PadéNet variants with moderate numerator and denominator orders combined with the $\mathrm{LeakyReLU}$ activation function exhibit the most competitive performance profiles, maintaining higher accuracy even in severely degraded noise environments.

For example, using Accelerometer-1 input, the \(P{=}1,Q{=}1\) configuration with $\mathrm{LeakyReLU}$ sustains accuracies exceeding \(93\%\) from as low as \(2\)~dB SNR, reaching \(98.91\%\) at \(10\)~dB. A similar pattern emerges for Accelerometer-3, where the \(P{=}1,Q{=}2\) $\mathrm{LeakyReLU}$ model surpasses \(90\%\) accuracy from \(0\)~dB onwards, achieving \(96.29\%\) at \(10\)~dB. The performance gap is wider in low-SNR regimes (\(-4\)~dB to \(0\)~dB), where accurate recognition is typically most challenging. On the acoustic modality at \(-4\)~dB, for instance, the \(P{=}1,Q{=}2\) $\mathrm{LeakyReLU}$ PadéNet attains \(52.07\%\) accuracy, representing an absolute gain of over \(6\%\) compared to the best-performing 1D CNN baseline (\(45.95\%\)). However, it should be noted that the acoustic modality is more affected by noise than the vibration modality, exhibiting greater performance loss under noisy conditions.

\begin{table*}[htbp]
\caption{Model complexity and inference performance.\label{tab_params}}
\centering
\small
\begin{tabularx}{\linewidth}{lccccc}
\toprule
\makecell{\textbf{Model} \\[-1pt] \scriptsize(config)} &
\makecell{\textbf{\# Params} \\[-1pt] \scriptsize(K)} &
\makecell{\textbf{FLOPs} \\[-1pt] \scriptsize(M)} &
\makecell{\textbf{Inf.\ Time} \\[-1pt] \scriptsize(ms)} &
\makecell{\textbf{Float32} \\[-1pt] \scriptsize(Model KB)} &
\makecell{\textbf{Quantized TFLite} \\[-1pt] \scriptsize(Model KB)} \\
\midrule
$P{=}1,\ Q{=}0$ (CNN)            &  58.38 & 14.79 & 3.159 $\pm$ 0.050 & 228.03 &  78.24 \\
$P{=}1,\ Q{=}1$ (PadéNet)        & 101.61 & 29.33 & 6.110 $\pm$ 0.048 & 396.91 & 127.08 \\
$P{=}1,\ Q{=}2$ (PadéNet)        & 144.84 & 43.97 & 9.591 $\pm$ 0.098 & 565.78 & 181.92 \\
$P{=}2,\ Q{=}0$ (Self-ONN)       & 101.83 & 29.49 & 5.542 $\pm$ 0.168 & 397.78 & 134.19 \\
$P{=}2,\ Q{=}1$ (PadéNet)        & 145.06 & 44.03 & 8.578 $\pm$ 0.224 & 566.66 & 182.58 \\
$P{=}3,\ Q{=}0$ (Self-ONN)       & 145.29 & 44.19 & 7.677 $\pm$ 0.127 & 567.53 & 186.85 \\
\bottomrule
\end{tabularx}
\end{table*}

While Self-ONNs demonstrate competitive behaviour at certain mid-SNR conditions, they generally lag behind PadéNets across both extreme and moderate noise levels. The consistently high accuracy observed for PadéNets at SNR values not explicitly encountered during training further indicates strong generalization to previously unseen noise conditions. These findings suggest that the PadéNet architecture offers enhanced noise robustness that remains consistent across different sensing modalities.

\subsection{Computational Complexity Analysis} \label{sec4_4}
The computational demands of the 1D CNN, 1D Self-ONN, and 1D PadéNet models were evaluated through their number of trainable parameters and average inference times, as summarized in Table~\ref{tab_params}. These metrics are critical for assessing the feasibility of deploying fault diagnosis models on resource-constrained devices, where computational efficiency must be balanced against diagnostic accuracy. For each configuration, the trainable parameter count, FLOPs (floating-point operations), inference duration per 1000-sample window measured with TensorFlow Lite on a Raspberry~Pi~4 Model~B (mean~$\pm$~s.d.\ over 100 runs), and the serialized model sizes for Float32 and 8-bit dynamic-range quantized TFLite exports are provided. The baseline 1D CNN ($P{=}1,Q{=}0$) requires the fewest parameters (58{,}376) and is the fastest at $3.159\pm0.050$\,ms per window, while the highest-accuracy PadéNet settings evaluated---$P{=}2,Q{=}1$ and $P{=}1,Q{=}2$---run in $8.578\pm0.224$\,ms and $9.591\pm0.098$\,ms, respectively. This corresponds to a throughput of roughly $100$--$320$ windows/s across all variants, indicating that the models satisfy real-time constraints on our edge platform. Model footprints span $228$--$568$\,KB for Float32 and $78$--$187$\,KB for the quantized TFLite models, which fit within the flash budgets of many Cortex-M--class MCUs. FLOPs are computed on frozen TensorFlow graphs using the TensorFlow profiler.

While increasing the degree of the numerator ($P$) and denominator ($Q$) generally enhances fault diagnosis accuracy, it also results in increased computational complexity for the 1D PadéNet models. Nevertheless, with moderate values of $P$ and $Q$, the increase in trainable parameters remains manageable, and the inference time stays within a few milliseconds, making 1D PadéNets a competitive option for fault diagnosis systems that can be deployed on the edge devices.

\section{Conclusions} \label{sec5}

This study proposed 1D Padé Approximant Neural Networks (PadéNets) for classifying mechanical and electrical faults in three-phase induction motors using vibration and acoustic data. By leveraging rational-function-based nonlinearities, PadéNets consistently outperformed traditional CNNs and Self-ONNs across all sensor inputs. The best performance was achieved on the first accelerometer with an average accuracy of $99.96\%$, while acoustic-based classification reached $98.33\%$, highlighting the potential of 1D PadéNets for both contact and non-contact condition monitoring. The models demonstrated strong generalization across sensor positions, with lower complexity and faster inference compared to state-of-the-art alternatives. These results establish PadéNets as effective and efficient tools for real-time fault diagnosis. Future work could focus on extending 1D PadéNets to variable-speed operating conditions and integrating multi-sensor data fusion to advance intelligent fault diagnosis in electrical machines.

\backmatter

\bmhead{Acknowledgements}
This work was supported by the Scientific and Technological Research Council of Turkey (TÜBİTAK).

\section*{Declarations}

\begin{itemize}
    \item Funding:
    The Scientific and Technological Research Council of Turkey (TÜBİTAK) supports the work of Sertac Kilickaya through the 2211-E National PhD Scholarship Program. 
    \item Conflict of interest/Competing interests:
    The authors declare no conflicts of interest.
    \item Data availability: 
    The original data presented in the study are openly available in Mendeley Data at \url{https://dx.doi.org/10.17632/msxs4vj48g.1}.
    \item Materials availability:
    Not applicable
    \item Code availability: 
    Not applicable
    \item Author contribution:
    Conceptualization, S.K. and L.E.; methodology, S.K. and L.E.; software, S.K.; validation, S.K.; writing---original draft preparation, S.K.; writing---review and editing, S.K. and L.E.; supervision, L.E. All authors have read and agreed to the published version of the manuscript.
\end{itemize}


\bibliography{sn-bibliography}

\end{document}